\documentclass{article}

\usepackage{microtype}
\usepackage{graphicx}
\usepackage{subfigure}
\usepackage{booktabs} % for professional tables

\usepackage[bookmarks=false,colorlinks]{hyperref}

\usepackage[accepted]{icml2024}

\usepackage{amsmath}
\usepackage{amssymb}
\usepackage{mathtools}
\usepackage{amsthm}

\usepackage[capitalize,noabbrev]{cleveref}
\usepackage{multirow}
\usepackage{amsmath,bm}
\usepackage{bbding}
\usepackage{amsmath}
\usepackage{amssymb}

\usepackage[labelfont = it]{caption}

\theoremstyle{plain}

\theoremstyle{definition}

\theoremstyle{remark}

\usepackage[textsize=tiny]{todonotes}

\usepackage{xcolor}

\usepackage{enumitem}

\icmltitlerunning{Addressing Long-Tail Noisy Label Learning Problems: a Two-Stage Solution with Label Refurbishment Considering Label Rarity}

\begin{document}

\twocolumn[
%%%%%%%%%%%%%%%%%%%%%%%%%%%%%%%%%%%%%%
%%%%%%%%%%%%%%%%%%%%%%%%%%%%%%%%%%%%%%
% \icmltitle{Enhancing Long-Tail Noisy Label Classification by Soft Label Refurbishing and Multi-Expert Ensemble Learning}
\icmltitle{Addressing Long-Tail Noisy Label Learning Problems: a Two-Stage Solution with Label Refurbishment Considering Label Rarity}

%Robust Long-Tail Noisy Label Classification}

%\icmltitle{Robust Long-Tail Noisy Label Classification via Soft Label Refurbishing and Noise-Tolerant Loss}

%"Soft Label Refurbishing for Robust Long-Tail Classification: A Novel Approach to Address Noisy Labels and Class Imbalance"
%"Contrastive Learning and Noise-Tolerant Loss for Improved Long-Tail Classification: A Soft Label Refurbishing Approach"
%"Enhancing Deep Learning Generalization through Soft Label Refurbishing: A Solution to Noisy Labels and Class Imbalance in Long-Tailed Datasets"
%"Addressing Noisy Labels and Class Imbalance in Long-Tail Datasets: A Soft Label Refurbishing Approach with Contrastive Learning"
%"Soft Label Refurbishing: Overcoming Noisy Labels and Class Imbalance for Improved Generalization in Long-Tail Classification"
%%%%%%%%%%%%%%%%%%%%%%%%%%%%%%%%%%%%%%
%%%%%%%%%%%%%%%%%%%%%%%%%%%%%%%%%%%%%%

% It is OKAY to include author information, even for blind
% submissions: the style file will automatically remove it for you
% unless you've provided the [accepted] option to the icml2024
% package.

% List of affiliations: The first argument should be a (short)
% identifier you will use later to specify author affiliations
% Academic affiliations should list Department, University, City, Region, Country
% Industry affiliations should list Company, City, Region, Country

% You can specify symbols, otherwise they are numbered in order.
% Ideally, you should not use this facility. Affiliations will be numbered
% in order of appearance and this is the preferred way.
\icmlsetsymbol{equal}{*}

\begin{icmlauthorlist}
\icmlauthor{Ying-Hsuan Wu }{nycu}
\icmlauthor{Jun-Wei Hsieh}{nycu}
\icmlauthor{Xin Li}{alb}
\icmlauthor{Shin-You Teng}{nycu}
\icmlauthor{Yi-Kuan Hsieh}{nycu}
\icmlauthor{Ming-Ching Chang}{alb}
% \icmlauthor{Firstname7 Lastname7}{comp}
%\icmlauthor{}{sch}
% \icmlauthor{Firstname8 Lastname8}{sch}
% \icmlauthor{Firstname8 Lastname8}{yyy,comp}
%\icmlauthor{}{sch}
%\icmlauthor{}{sch}
\end{icmlauthorlist}

\icmlaffiliation{nycu}{College of Artificial Intelligence and Green Energy, National Yang Ming Chiao Tung University, Taiwan}
\icmlaffiliation{alb}{Department of Computer Science, University at Albany, SUNY, NY, USA}
% \icmlaffiliation{sch}{School of ZZZ, Institute of WWW, Location, Country}

\icmlcorrespondingauthor{Ying-Hsuan Wu}{vongola3088.ai10@nycu.edu.tw}
\icmlcorrespondingauthor{Jun-Wei Hsieh}{jwhsieh@nycu.edu.tw}
\icmlcorrespondingauthor{Xin Li}{xli48@albany.edu}
\icmlcorrespondingauthor{Ming-Ching Chang}{mchang2@albany.edu}

% You may provide any keywords that you
% find helpful for describing your paper; these are used to populate
% the "keywords" metadata in the PDF but will not be shown in the document
\icmlkeywords{Machine Learning, ICML}

\vskip 0.3in
]

% this must go after the closing bracket ] following \twocolumn[ ...

% This command actually creates the footnote in the first column
% listing the affiliations and the copyright notice.
% The command takes one argument, which is text to display at the start of the footnote.
% The \icmlEqualContribution command is standard text for equal contribution.
% Remove it (just {}) if you do not need this facility.

\printAffiliationsAndNotice{}  % leave blank if no need to mention equal contribution
% \printAffiliationsAndNotice{\icmlEqualContribution} % otherwise use the standard text.

%%%%%%%%%%%%%%%%%%%%%%%%%%%%%%%%%%%%%%
%%%%%%%%%%%%%%%%%%%%%%%%%%%%%%%%%%%%%%
\begin{abstract}
Real-world datasets commonly exhibit noisy labels and class imbalance, such as long-tailed distributions. While previous research addresses this issue by differentiating noisy and clean samples, reliance on information from predictions based on noisy long-tailed data introduces potential errors.  To overcome the limitations of prior works, we introduce an effective two-stage approach by combining soft-label refurbishing with multi-expert ensemble learning. In the first stage of robust soft label refurbishing, we acquire unbiased features through contrastive learning, making preliminary predictions using a classifier trained with a carefully designed BAlanced Noise-tolerant Cross-entropy (BANC) loss. In the second stage, our label refurbishment method is applied to obtain soft labels for multi-expert ensemble learning, providing a principled solution to the long-tail noisy label problem. Experiments conducted across multiple benchmarks validate the superiority of our approach, Label Refurbishment considering Label Rarity ($LR^2$), achieving remarkable accuracies of 94.19\% and 77.05\% on simulated noisy CIFAR-10 and CIFAR-100 long-tail datasets, as well as 77.74\% and 81.40\% on real-noise long-tail datasets, Food-101N and Animal-10N, surpassing existing state-of-the-art methods.

%We present a novel approach to address long-tail noisy label classification using soft label refurbishing and contrastive learning. Real-world datasets often suffer from both noisy labels and class imbalance, both of which are critical factors leading to poor generalization of deep learning models. Previous research has developed methods to tackle this problem, mainly focusing on differentiating between noisy samples and clean samples from the tail class, but the information used for discrimination often comes from deep networks trained on noisy long-tailed data, leading to potential errors in judgment. In this paper, to address this problem and overcome the limitations of previous work, we propose an approach based on soft label refurbishing. Specifically, we first obtains unbiased and robust features through contrastive learning and makes preliminary predictions using a classifier trained with the proposed noise-tolerant loss. Then, the proposed label refurbishment method is utilized to obtain soft labels for multi-expert ensemble learning, further addressing the long-tail problem. Experiments on multiple benchmarks have demonstrated the superiority of our approach.
\end{abstract}

%%%%%%%%%%%%%%%%%%%%%%%%%%%%%%%%%%%%%%
%%%%%%%%%%%%%%%%%%%%%%%%%%%%%%%%%%%%%%
%%%%%%%%%%%%%%%%%%%%%%%%%%%%%%%%%%%%%%
\vspace{-4mm}
\section{Introduction}

Deep learning has experienced notable advancements in various domains, significantly driven by the advantages of large-scale, meticulously annotated datasets~\cite{krizhevsky2012imagenet, ren2015faster, devlin2018bert}. However, the practical acquisition of such large-scale datasets is inherently challenging, primarily due to two factors. First, the complexity and uncertainty of the data sources, combined with the potential errors arising from the human annotation process, contribute to the imperfections~\cite{xiao2015learning, li2019learning}. Second, datasets commonly exhibit a long-tailed distribution, resulting in a disparity in sample numbers between different categories~\cite{lee2021learning, zhang2023deep}. In particular, these dual imperfections often coexist, further complicating the data landscape in real-world scenarios.

%-------------------------------------
\begin{figure}[t]
%\vskip 0.2in
\centerline{
\includegraphics[width=\columnwidth, height = 0.6\columnwidth]{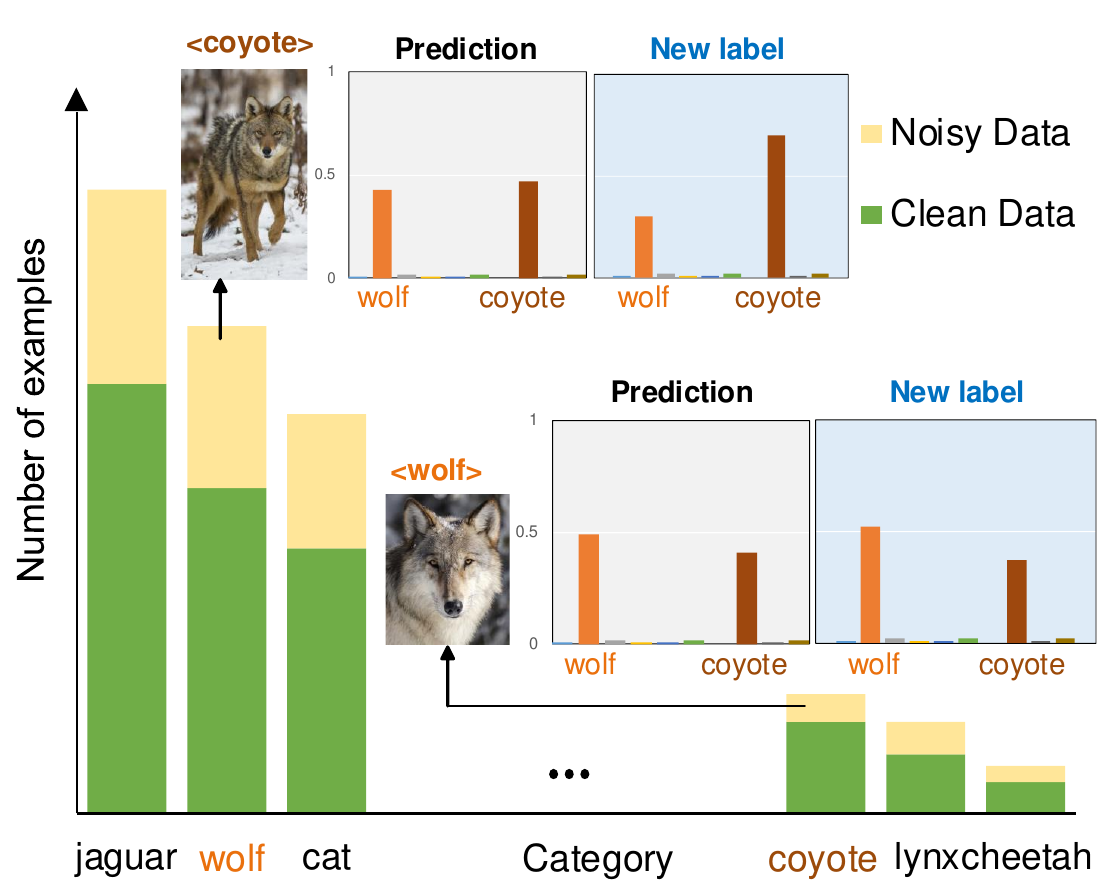}
\vspace{-2.5mm}
}
\caption{A fundamental challenge in long-tailed noisy label classification is the potential confusion between noisy labels in head classes and clean labels in tail classes. Data rarity in tail classes further complicates the label refurbishment in these instances.}
\label{fig:teaser}
\vspace{-4mm}
\end{figure}
%-------------------------------------

Numerous studies have addressed challenges associated with noisy labels and long-tail data independently, but their effectiveness diminishes when both conditions coexist. Methods designed for noisy labels often assume a class-balanced dataset, neglecting rare classes in long-tailed distributions and increasing susceptibility to noisy labels~\cite{song2022learning}. Additionally, similar training dynamics between clean data from tail classes and data with incorrect labels may result in an erroneous correction of tail class samples~\cite{cao2020heteroskedastic, xia2021sample}, exacerbating the long-tail problem. Conversely, methods exclusively designed for long-tailed data may use re-sampling or re-weighting techniques to balance the classifier~\cite{estabrooks2004multiple, ren2020balanced}. However, the coexistence of noisy labels can compromise these efforts, causing the model to learn incorrect features on tail-class samples.

%-------------------------------------
\begin{figure*}[t]
\centerline{
\includegraphics[width=\textwidth, height=0.27\textwidth]{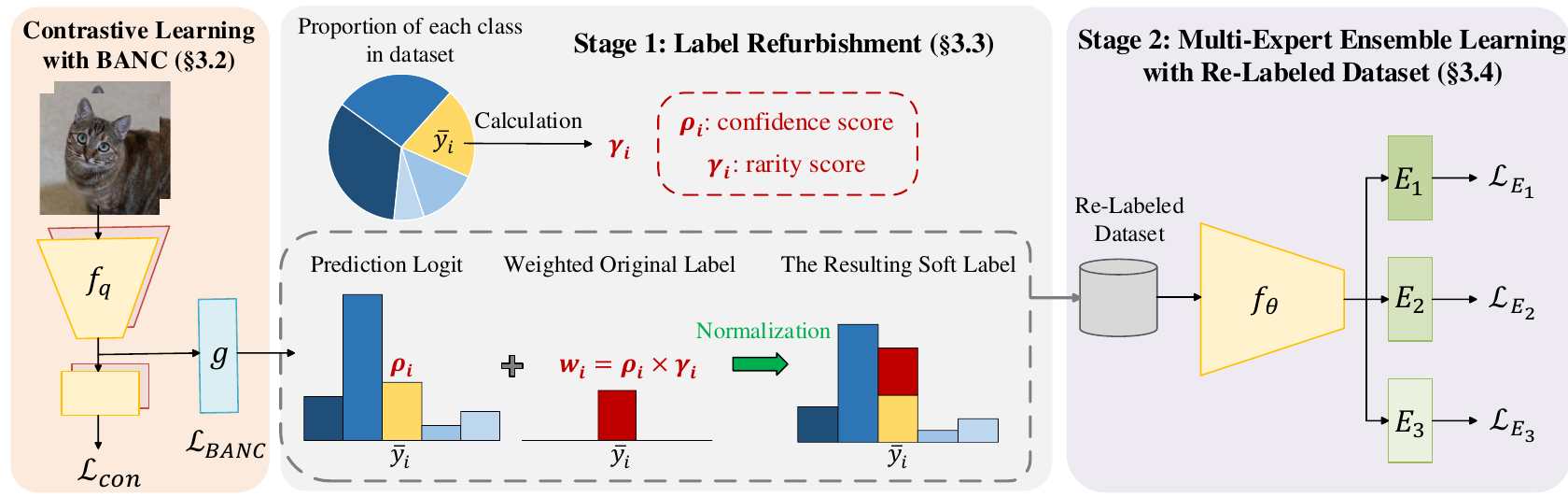}
\vspace{-2mm}
}
\caption{The proposed approach tackles the long-tail noisy label learning problem through a two-stage process. Stage 1 involves initial prediction using contrastive learning using a newly designed BAlanced Noise-tolerant Cross entropy (BANC) loss ($\S$~\ref{section:contrastive}), followed by label refurbishment ($\S$~\ref{section:re-label}). Stage 2 employs ensemble learning using three expert modules ($\S$~\ref{section:ensemble}), specifically designed to enhance long-tail classification.
}
\vspace{-4mm}
\label{fig:architecture}
\end{figure*}
%-------------------------------------

The specialized methods discussed above struggle to achieve robust and stable performance in handling datasets containing both noisy labels and long-tail distributions. Existing methods mainly focus on distinguishing between noisy and clean samples from tail classes for subsequent processing~\cite{wei2021robust, yi2022identifying, lu2023label}. However, these methods often rely on information from a pre-screening network trained on long-tailed data with noise, potentially leading to erroneous judgments. For example, as shown in Fig.~\ref{fig:teaser}, the distribution of noisy labels in head-class might appear similar to that of clean labels of tail-class. This observation motivates a new approach: instead of rigidly determining whether a sample is mislabeled, we propose to consider the class distribution, original labels, and information from the pre-screening network, to perform a soft label refurbishment on the dataset. Moreover, the ensemble trick \cite{du2023no} has shown potential in handling a long-tailed distribution. Therefore, we conjecture that a holistic perspective can offer a more flexible and robust solution to the challenge of long-tailed noisy labels.

This paper tackles the dual challenges of noisy labeling and long-tailed distribution in datasets through a two-stage training strategy, focusing on effective soft label refurbishment. Fig.~\ref{fig:architecture} provides an overview of our method.  In the first stage, unsupervised contrastive learning is employed to generate representations for all training instances. Since the process of representation learning is not affected by the imbalance in class distribution and the corruption of training labels, the representations obtained are inherently unbiased and robust~\cite{liu2021self, ghosh2021contrastive}. Building upon these representations, we introduce a novel loss design to train a balanced and noise-tolerant classifier for data pre-screening. Following the pre-screening phase, the soft label refurbishment process begins. By combining original labels with weighted adjustments that consider both prediction confidence and class rarity, we generate robust soft labels for each sample. Notably, this method does not solely depend on pre-screening predictions to identify mislabeled samples, nor does it directly replace labels based on predicted classes. Instead, it conditionally integrates the prediction results with the original labels, allowing for a soft update to the dataset labels.

Following label refurbishment, where the class distribution in the dataset remains imbalanced, we address the long-tail problem in the second stage. Utilizing the unbiased and robust representations obtained in the initial stage, we employ ensemble learning to train multiple expert classifiers, each specialized in recognizing different categories such as many-shot, medium-shot, and few-shot categories. Soft labels generated through a well-designed strategy guide training in this stage, enhancing the model robustness to long-tailed distributions, as well as generalization capability. Notably, for experts specializing in medium and few-shot categories, we adopt soft label-based class distribution statistics for loss weighting. Compared to the conventional use of hard labels for class distribution statistics~\cite{ren2020balanced, hong2021disentangling}, our approach is more in line with the real class distribution, particularly in cases where the original labels are corrupted. Consequently, it enhances the precision of expert loss weighting and contributes to improved model performance.
The main contributions of our work are summarized as follows:
\begin{itemize}[leftmargin=10pt] \itemsep -.1em
\item We address the practical challenge of robust classification from datasets that simultaneously exhibit both noisy label and long-tailed distributions. 
\item Our solution is a two-stage model training strategy. We handle noisy labels by soft label refurbishment, which is performed based on training a sample pre-screening model via unsupervised contrastive learning. The long-tail issue is then addressed by using three expert classifiers specialized in many-shot, median-shot, and few-shot (rare label) categories, respectively. 
%
%\item We propose an advanced training strategy. Through balanced and noise-tolerant initial predictions, coupled with information about the dataset's class distribution, we generate soft labels for ensemble learning. This strategy effectively addresses the impact of both noisy labels and long-tailed distribution, enhancing the model's robustness.
%
\item We conducted extensive experiments on both simulated and real-world datasets. The results show that our approach, label refurbishment considering label rarity ($LR^2$), outperforms existing state-of-the-art, achieving remarkable accuracies of 94.19\% and 77.05\% on simulated noisy long-tailed datasets of CIFAR-10 and CIFAR-100, as well as 77.74\% and 81.40\% on real-noise long-tailed datasets, Food-101N and Animal-10N, respectively.
\end{itemize}

%%%%%%%%%%%%%%%%%%%%%%%%%%%%%%%%%%%%%%
%%%%%%%%%%%%%%%%%%%%%%%%%%%%%%%%%%%%%%
%%%%%%%%%%%%%%%%%%%%%%%%%%%%%%%%%%%%%%
\section{Related Work}

%%%%%%%%%%%%%%%%%%%%%%%%%%%%%%%%%%%%%%
%%%%%%%%%%%%%%%%%%%%%%%%%%%%%%%%%%%%%%
\subsection{Label-Noise Learning}

One direct strategy to deal with noisy data is to separate noisy samples from clean ones through sample selection, reducing the proportion of noisy samples in the training set. Methods such as co-teaching~\cite{han2018co} and DivideMix~\cite{li2020dividemix} use small-loss techniques, while Jo-SRC~\cite{yao2021jo} and UNICON~\cite{karim2022unicon} employ Jensen-Shannon divergence for sample selection. Another strategy to address this issue involves correcting labels to prevent model overfitting to incorrectly labeled instances. SELFIE~\cite{song2019selfie} discerns samples with consistently accurate label predictions, identifying them as refurbishable samples. Only the labels of these refurbishable instances are corrected, thereby minimizing the occurrence of erroneously rectified cases. SEAL~\cite{chen2021beyond} employs a method that calculates the average softmax outputs of a deep neural network (DNN) for each example over the entire training process. Subsequently, it retrains the DNN using the averaged soft labels.
Other methods include adjusting the loss function to mitigate the impact of noisy data~\cite{patrini2017making,hendrycks2018using,shu2019meta}, and employing regularization constraints to prevent the model from overfitting to noisy data~\cite{pereyra2017regularizing,xia2020robust}. Most recently, a Robust Label Refurbishment (Robust LR) method was developed in \cite{chen2023two}, which integrates pseudo-labeling and confidence estimation techniques to refurbish noisy labels.

%%%%%%%%%%%%%%%%%%%%%%%%%%%%%%%%%%%%%%
%%%%%%%%%%%%%%%%%%%%%%%%%%%%%%%%%%%%%%
\subsection{Long-tail Learning}

Existing methods tackling long-tail data primarily focus on: (1) re-balancing data distributions, such as resampling~\cite{haixiang2017learning,ma2018dimensionality,byrd2019effect}, and data augmentation~\cite{chou2020remix,zang2021fasa,zhong2021improving}; (2) re-designing loss function, developing long-tail robust loss functions to enhance model generalization~\cite{cao2019learning,cui2019class,ren2020balanced}; (3) decoupling representation learning and classifier learning to make the obtained feature extractor less susceptible to the effects of long-tail distribution~\cite{kang2019decoupling,zhong2021improving}; (4) utilizing transfer learning techniques to transfer knowledge from head classes to tail classes~\cite{wang2020long,he2021distilling}. A recent method, local and global logit adjustment (LGLA)~\cite{tao2023local}, trains expert models with complete data that spans all classes. This method enhances the differentiation  among these models by meticulous logit adjustments. 

%%%%%%%%%%%%%%%%%%%%%%%%%%%%%%%%%%%%%%
%%%%%%%%%%%%%%%%%%%%%%%%%%%%%%%%%%%%%%
\subsection{Label-Noise Learning on Long-tailed Data}

RoLT~\cite{wei2021robust} shows that techniques selecting only samples with small loss for training fails under long-tail label distribution. The authors of RoLT propose a prototype noise detection method to better differentiate mislabeled examples from rare instances. An adaptive method is introduced in HAR~\cite{cao2020heteroskedastic} to regularize noise and tail class samples. ULC~\cite{huang2022uncertainty} improves the separability of noisy samples by introducing uncertainty. H2E~\cite{yi2022identifying} reduces hard noise to easy noise by learning a classifier as a noise identifier that is invariant to variations in the class and context distribution. TABASCO~\cite{lu2023label} uses two complementary separation metrics to differentiate between clean and noisy samples, especially for tail classes. A representation calibration method called RCAL was proposed in \cite{zhang2023noisy} to recover the underlying representation distributions from mislabeled and class-imbalanced data.

%%%%%%%%%%%%%%%%%%%%%%%%%%%%%%%%%%%%%%
%%%%%%%%%%%%%%%%%%%%%%%%%%%%%%%%%%%%%%
%%%%%%%%%%%%%%%%%%%%%%%%%%%%%%%%%%%%%%
\section{Methodology}

In this paper, we propose a two-stage training strategy to address the learning problem that involves both long-tailed distribution and label noise simultaneously. %Generally, our approach consists of two stages: (1) enhancing representations through contrastive learning, making preliminary predictions using the proposed loss-trained classifier (Sec.~\ref{section:contrastive}), and then obtaining soft labels for samples through the introduced label refurbishment method (Sec.~\ref{section:re-label}); (2) utilizing soft labels for multi-expert ensemble learning (Sec.~\ref{section:ensemble}). 
The overall framework is shown in Fig.~\ref{fig:architecture}. %In the following part, we first introduce the problem definition and then provide detailed explanations of the proposed training strategy techniques.

%%%%%%%%%%%%%%%%%%%%%%%%%%%%%%%%%%%%%%
%%%%%%%%%%%%%%%%%%%%%%%%%%%%%%%%%%%%%%
\subsection{Problem Definition}

In this paper, scalars are in lowercase letter, and vectors are in lowercase boldface letters. Given an imbalanced and noisily labeled training dataset $\bar{\mathcal{D}}=\{(\bm{x}_i,\bar{y_i})\}^{N}_{i=1}$, $\bm{x}_i$ denotes the $i$-th instance and its label $\bar{y_i}\in[K]$ may be incorrect, where $N$ is training sample size, and $K$ is the number of classes. For label $\bar{y_i}$, the corresponding true label is denoted by $y_i$, which is unobservable. Let $n_k$ denote the number of training data belonging to the $k$-th class. Without loss of generality, we assume that the classes are sorted in descending order based on the number of training samples for each class, $i.e., n_1 \ge ... \ge n_K$. Our objective is to train a robust classifier using only the imbalanced training dataset with noisy labels, enabling the classifier to accurately infer the labels for the unknown instances.

%%%%%%%%%%%%%%%%%%%%%%%%%%%%%%%%%%%%%%
%%%%%%%%%%%%%%%%%%%%%%%%%%%%%%%%%%%%%%
\subsection{Robust Initial Predictions: Contrastive Learning with BAlanced Noise-tolerant Cross entropy}  \label{section:contrastive}

\textbf{Contrastive learning.} 
In the first stage, we enhance robust data representations for instances with both noisy labels and long-tailed distributions using self-supervised contrastive learning. This approach does not access to the original training labels, ensuring that the learned representations are not influenced by incorrect labels~\cite{ghosh2021contrastive}. 
Additionally, prior research \cite{liu2021self} has shown that contrastive learning can mitigate the impact of long-tailed data on the network. Consequently, we leverage contrastive learning to improve representations in the presence of both noisy labels and long-tailed distributions.

% In more detail, we utilized the Siamese network setup from MoCo \cite{chen2020improved}. MoCo framework consists of two networks with the same structure, $i.e.$, \textit{query network} and \textit{key network}. For each network, it usually contains one encoder CNN and one two-layer MLP transform. For each input image, we applied two random augmentations, thereby generating two views. In a batch of two-viewed images denoted as $\mathcal{B}=(\mathcal{B}_{v1}, \mathcal{B}_{v2})$, where $\mathcal{B}_{v1}$ and $\mathcal{B}_{v2}$ are then fed into the query network and the key network, producing representations $\mathcal{Z}_{v1}$ and $\mathcal{Z}_{v2}$. MoCo also includes a large queue $Q$ to store a substantial number of sample features, contributing to the learning of well-structured representations. Unlike the original setup of MoCo, we omitted the momentum update between the key encoder and the query encoder, meaning we froze the parameters of the query encoder and only updated the gradients for the key encoder. Recent research \cite{chen2021exploring} has demonstrated that halting the gradient updates for the query encoder contributes to improving model performance and robustness. The contrastive loss for the $i$-th input $\bm{x}_i$ can be expressed as:

In more detail, we utilized the MoCo~\cite{chen2020improved} network setup for contrastive learning. The MoCo framework comprises two networks with the same structure, namely the \textit{query network} and \textit{key network}. Each network includes a typical CNN encoder and a two-layer MLP. For the $i$-th input $\bm{x}_i$, two random augmentations are applied, resulting in two views $\bm{x}_i^q$ and $\bm{x}_i^k$. These views are then input into the query network and the key network, generating representations $\bm{z}_i^q$ and $\bm{z}_i^k$. MoCo also integrates a large queue $Q$ to store a substantial number of sample features, aiding in the learning of well-structured representations.
Unlike the original MoCo setup, we excluded the momentum update between the key encoder and the query encoder, effectively freezing the parameters of the query encoder and updating only the gradients for the key encoder. Recently, \cite{chen2021exploring} show that halting gradient updates for the query encoder enhances model performance and robustness.

Let $\mathcal{A}(i)$ denote the union of representations from $Q$ and other samples except $\bm{x}_i$ in the current batch.
By minimizing the following contrastive loss, improved representations for the input $\bm{x}_i$ can be achieved:
\vspace{-2mm}
\begin{equation}
\mathcal{L}_{con}(\bm{x}_i)=-\log\frac{\exp(\bm{z}_i^q\cdot\bm{z}_i^k/\tau)}{{\sum_{\bm{z}_j\in \mathcal{A}(i)}}\exp(\bm{z}_i^q\cdot{\bm{z}_j}/\tau)},
\label{eq:contrastive}
\end{equation}
\vskip -2mm
% \[\mathcal{A}(i)=\{{\bm{z}_j}\in\mathcal{Q}\cup\mathcal{Z}_{v1}\cup\mathcal{Z}_{v2}\}\setminus\{{\bm{z}_j}\in\mathcal{Z}_{v1}:j=i\},\]
where $\tau>0$ is a temperature parameter. 

%%%%%%%%%%%%%%%%%%%%%%%%%%%%%%%%%%%%%%
\begin{figure*}[t]
\centerline{
\includegraphics[width=0.73\linewidth, height=0.18\linewidth]{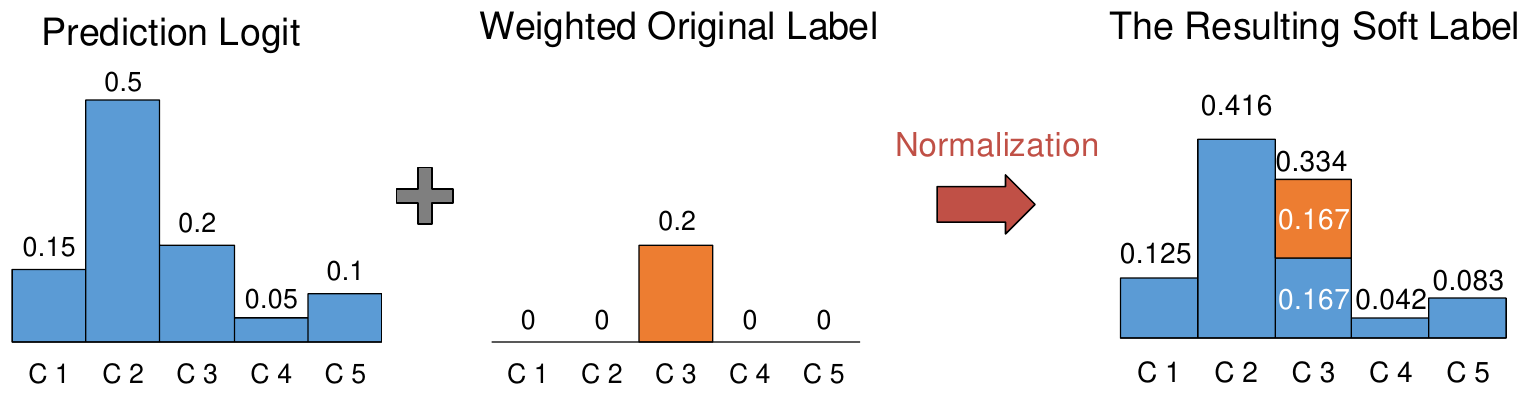}
\vspace{-4mm}
}
\caption{Label refurbishment involves the use of soft labels determined by the confidence of the original labels. In this example, the prediction score of C3 is updated after normalization.}
\vspace{-4mm}
\label{fig:relabel}
\end{figure*}
%%%%%%%%%%%%%%%%%%%%%%%%%%%%%%%%%%%%%%

\textbf{Balanced noise-tolerant cross entropy.} 
Having obtained features resilient to label influence through self-supervised contrastive learning, the next step involves employing a robust classifier to correct the potentially erroneous labels in the class imbalance conditions. Drawing inspiration from Symmetric Cross Entropy (SCE)~\cite{wang2019symmetric}, we introduce a novel loss function named BAlanced Noise-tolerant Cross-entropy (BANC), designed for enhanced noise resilience and balance effects. Specifically, SCE, based on Cross Entropy, introduces a symmetric term to facilitate the learning of label-to-prediction distribution. To provide a detailed calculation, let $\bm{v}_i=f_q(\bm{x}_i)$ that is extracted from the query encoder $f_q$ for subsequent classification. In addition, the one-hot representation of $\bar{y}_i$ is denoted as $\bm{\bar{y}}_i$ = $[\bar{y}_i^1, \dots, \bar{y}_i^k, \dots, {\bar{y}_i^K}]$.
Moreover, we use $g(\cdot)$ to represent the classifier, and the predicted logit $g(\bm{v}_i)=[{p_i^1}, \dots, p_i^k, \dots, {p_i^K}]$. SCE is then formulated as:
\vspace{-2mm}
\begin{equation} 
\mathcal{L}_{SCE}(\bm{x}_i)= - \sum\limits_{k = 1}^K  {\bar y_i^k\log \bar{p}_i^k - \sum\limits_{k = 1}^K \log (\bar y_i^k)\bar{p}_i^k} ,
\end{equation}
\vskip -2mm
where $\bar{p}_i^k$ denotes the value of $p_i^k$ after the Softmax normalization. 
%Since the value of $\bar{y}_i^k$ is either one or zero, $\log(\bar{y}_i^k)$ is not defined if $\bar{y}_i^k=0$. 
To avoid the specific handling of the undefined value of $\log(0)$ in SCE, we replace $-\log(\bar{y}_i^k)$ with a linear function $c\,(1-\bar{y}_i^k)$, where $c>0$ and serves as a scaling coefficient. We define Our BANC loss as:
\vspace{-2mm}
\begin{equation} 
\mathcal{L}_{BANC}(\bm{x}_i)= - \sum\limits_{k = 1}^K  {\bar y_i^k\log \bar{p}_i^k + \sum\limits_{k = 1}^K c (1-\bar{y}_i)\bar{p}_i^k} ,
\end{equation}
\vskip -2mm
The BANC loss introduces label error tolerance in the loss computation for the following reasons. When $c>0$ and $\bar{y}_i^k=0$, the impact of $c\,(1-\bar{y}_i^k)$ is to impose a relatively smaller penalty $c$ to the increase in loss under label errors, effectively mitigating the adverse effects of label noise on the model. 
Similarly, in cases of class imbalance, this penalty enables the model to prioritize the prediction of minority classes, adjusting the loss by $c$. This adjustment helps address class imbalance issues, fostering a more balanced learning experience across different classes. Additional experiments in Appendix~\ref{appendix:sce} validate these arguments.

In summary, the loss used for prediction in the first stage comprises the contrastive loss $\mathcal{L}_{con}$ for contrastive learning and the BANC loss $\mathcal{L}_{BANC}$ for improving classifier training. Then, the overall loss for the first stage $\mathcal{L}_{S1}$ is:
\vspace{-2mm}
\begin{equation} 
\mathcal{L}_{S1}= (1-\alpha)\mathcal{L}_{con}+\alpha\mathcal{L}_{BANC},
\end{equation}
\vskip -2mm
where $\alpha$ is a hyperparameter to weight the contribution of the two loss terms, with default $0.2$. The influence of $\alpha$ is analyzed in Appendix~\ref{appendix:alpha}.

%%%%%%%%%%%%%%%%%%%%%%%%%%%%%%%%%%%%%%
%%%%%%%%%%%%%%%%%%%%%%%%%%%%%%%%%%%%%%
\subsection{Stage 1: Label Refurbishment} 
\label{section:re-label}

Following the initial label predictions, we employ a label refurbishment strategy that takes into account the challenges posed by noisy labels and long-tailed distributions in the training data, even with the contrastive learning and robust loss. The strategy proceeds as follows: If the predicted class aligns with the original label, we consider the original label likely correct and retain it. On the contrary, if the predicted class differs from the original label, we generate the new label by adding the predicted logit to the weighted original label and normalizing the result, as illustrated in Fig.~\ref{fig:relabel}. Let $l_i$ denote the predicted class of input $x_i$; that is, $l_i = \arg \mathop {\max }\limits_k p_i^k$. The label refurbishment is performed as:
\vspace{-2mm}
\begin{equation}
\bm{\hat{y}}_i
=
\begin{cases}
\bar{\bm{s}}_i, &\text{if}\;{l_i}\not=\bar{y}_i,\\
\bm{\bar{y}}_i, &\text{otherwise},
\end{cases}
\end{equation}
\vspace{-2mm}
\begin{equation}
\bar{\bm{s}}_i(k)
=
\frac{\bm{s}_i(k)}{\sum\limits_{k = 1}^K \bm{s}_i(k)},\quad\bm{s}_i(k)=p_i^k+w_i{{\bar{y}}_i^k},\\
\end{equation}
\vskip -2mm
where $w_i$ is the weight to weigh the original label of $\bm{x}_i$. The estimate of $w_i$ will be explained next.

We adhere to two principles when assigning the weight $w_i$ to the original labels: (1) Lower confidence in the predicted class indicates a higher likelihood of labeling error, resulting in a smaller weight assigned to the original label. (2) If the original label corresponds to a rare class, indicating a higher probability of prediction error, and considering the value of rare class samples, a larger weight is assigned to the original label. Specifically, we calculate the label weight $w_i$ as the multiplication of a confidence score and a rarity score estimate.
Specifically, let $\rho _i$ denote the predicted confidence of label $\bar{y}_i$, and $\gamma_i $ denote the rarity of class $\bar{y}_i$, then: 
\vspace{-2mm}
\begin{equation}
w_i = \rho_i \times \gamma_i,
\end{equation}
\vskip -2mm
where $\rho_i= {p_i^{\bar{y}_i}}$. The rarity score $\gamma_i$ is estimated as a function inversely proportional to the proportion of class $\bar{y}_i$ in the dataset. Let $n_{\bar{y}_i}$ denote the number of samples belonging to class $\bar{y}_i$, and let the proportion of class $\bar{y}_i$ be $h_i=\frac{n_{\bar{y}_i}}{N}$. When $h_i$ is close to zero, $\gamma_i$ approaches the maximum value 1 and when $h_i$ is greater than 0.5, $\gamma_i$ quickly decays to zero.  Then, a normal distribution function with zero mean is used to model $\gamma_i$ as the following (see Fig.~\ref{fig:ri} in Appendix~\ref{appendix:ri}): 
\vspace{-2mm}
\begin{equation}
\label{eq:ri}
\gamma_i = \exp\left( -{\frac{h_i^2}{\sigma^2}}\right),
\end{equation}
\vskip -2mm
where the variance $\sigma =0.2 $ by default. Through the proposed label-refurbishment strategy, samples with labels inconsistent with the classifier's predicted class receive new soft labels instead of being discarded or directly assigned to the predicted class. This helps avoid exacerbating the potential risks associated with the long tail and error correction.

%%%%%%%%%%%%%%%%%%%%%%%%%%%%%%%%%%%%%%
%%%%%%%%%%%%%%%%%%%%%%%%%%%%%%%%%%%%%%
\subsection{Stage 2: Multi-Expert Ensemble Learning} \label{section:ensemble}

After acquiring the newly estimated labels, we proceed to the second stage of model training that can further address the long-tail challenges. Specifically, we implement {\em ensemble learning} by constructing a three-expert model with two components: (1) a shared backbone $f_\theta$; and (2) three standalone expert classifiers $E_1$, $E_2$, and $E_3$, as illustrated in Fig.~\ref{fig:architecture}.
Note that the query encoder $f_q$ in the backbone, which is trained with contrastive learning in the first stage, remains uninfluenced by these labels, enabling it to learn unbiased and robust features.
Considering this, we configure the backbone $f_\theta$ to have identical parameter weights as $f_q$ and freeze $f_\theta$ to prevent gradient interference from expert training losses. 

To enhance model diversity, each expert is trained to specialize in different class groups. Specifically, the many-shot expert $E_1$ excels in recognizing main classes, the medium-shot expert $E_2$ focuses on balancing predictions across all classes, achieving a certain level of overall accuracy, and the few-shot expert $E_3$ is dedicated to identifying rare classes (and also compensate for the weakness of expert $E_1$). The following describes the three {\em shot-adaptive losses}, each tailored to suit the training of respective expert model.

\textbf{The many-shot expert $E_1$}:
Softmax cross-entropy loss is used to train this expert. Due to the long-tail distribution, the many-shot class tends to dominate the gradients, leading the expert predictions to bias towards the many-shot class naturally.
Similar to the definition of $g(\bm{v}_i)$, we use $g_{E_1}(\bm{v}_i)$ to denote the predicted result of $\bm{v}_i$ using expert $E_1$. In addition, we use ${g}_{E_1}^k(\bm{v}_i)$ to indicate the $k$-th entry of $g_{E_1}(\bm{v}_i)$, and $\hat{y}_i^k$ to indicate the $k$-th entry of $\bm{\hat{y}}_i$. Let $\Phi(\cdot)$ denote a Softmax normalization. The loss of the expert $E_1$ is:
\vspace{-3mm}
\begin{equation} 
\mathcal{L}_{E_1}(\bm{x}_i)= - \sum\limits_{k = 1}^K  {\hat y_i^k\log \Phi({g}_{E_1}^k(\bm{v}_i))}.
\vspace{-2mm}
\end{equation}
%\textred{The calculation of $n_k$ is discussed in detail in Section 3.5.}

\textbf{The medium-shot expert $E_2$}:
We train this expert using a loss similar to balanced softmax \cite{ren2020balanced}, where the loss is weighted based on the proportion of each class in the dataset. By balancing the class proportions in the gradient, this approach ensures that the expert performs well across all classes. Let $\bm{n}=[{n_1}, {n_2}, \dots, {n_K}]$ represent the sample size for each class. The loss of expert $E_2$ is:
\vspace{-2mm}
\begin{equation} 
\mathcal{L}_{E_2}(\bm{x}_i)= - \sum\limits_{k = 1}^K  {\hat y_i^k\log \Phi
\left(
{g}_{E_2}^k(\bm{v}_i)+\log n_k
\right)
}.
\vspace{-2mm}
\end{equation}

\textbf{The few-shot expert $E_3$}:
This expert specializes in identifying rare classes to address the challenge of limited data from few-shot classes. Drawing inspiration from the diversity loss~\cite{zhao2023mdcs}, we increase the loss weight, encouraging the expert's predictions to prioritize few-shot classes and consequently enhancing the recall rate for such classes. The loss of the expert $E_3$ is:
\vspace{-2mm}
\begin{equation} 
\mathcal{L}_{E_3}(\bm{x}_i)= - \sum\limits_{k = 1}^K  {\hat y_i^k\log \Phi
\left(
{g}_{E_3}^k(\bm{v}_i)+\log n_k^2
\right)
}.
\vspace{-2mm}
\end{equation}

\textbf{Training hard samples with soft labels.}
We use the newly obtained labels $\bm{\hat{y}}_i$ as ground truth for the training of hard samples, which includes the samples from minority classes and with noisy labels. Optimizing hard samples through soft labels enhances efficiency, prevents saturation, and promotes better learning, leading to improved robustness and generalization.

%%%%%%%%%%%%%%%%%%%%%%%%%%%%%%%%%%%%%%
\begin{table*}[t]
\caption{Test accuracy (\%) on simulated CIFAR-10 and CIFAR-100 with {\bf symmetric} noise. Bold indicates the highest score.\vspace{-2mm}
}
%\captionsetup{justification=centering, font={tiny}}
\label{table:sym}
%\vskip 0.15in
\begin{center}
\begin{small}
% \begin{sc}
\begin{tabular}{l|l|cccc|cccc}
\toprule
\multicolumn{2}{c}{Imbalance Ratio}& \multicolumn{4}{|c}{10} & \multicolumn{4}{|c}{100}\\ \midrule
\multicolumn{2}{c}{Dataset} & \multicolumn{2}{|c}{CIFAR-10} & \multicolumn{2}{c}{CIFAR-100} & \multicolumn{2}{|c}{CIFAR-10} & \multicolumn{2}{c}{CIFAR-100}\\ \midrule
\multicolumn{2}{c|}{Noise Rate \textbf{(Sym.)}}& 0.4           & 0.6          & 0.4        & 0.6  & 0.4           & 0.6          & 0.4        & 0.6\\ \midrule
Baseline             & CE          & 71.67         & 61.16        & 34.53      & 23.63   & 47.81         & 28.04        & 21.99      & 15.51\\ \midrule
\multirow{3}{*}{LT}    & LA          & 70.56         & 54.92        & 29.07      & 23.21 & 42.63         & 36.37        & 21.54      & 13.14 \\
                       & LDAM        & 70.53         & 61.97        & 31.30      & 23.13 & 45.52         & 35.29        & 18.81      & 12.65\\
                       & IB         & 73.24             & 62.62            & 32.40          & 25.84        & 49.07        & 32.54            & 20.34          & 12.10         \\ \midrule
\multirow{2}{*}{NL}    & DivideMix   & 82.67         & 80.17        & 54.71      & 44.98     & 32.42         & 34.73        & 36.20      & 26.29                \\
                       & UNICON      & 84.25         & 82.29        & 52.34      & 45.87    & 61.23         & 54.69        & 32.09      & 24.82                 \\ \midrule
\multirow{5}{*}{NL-LT}  & HAR         & 77.44         & 63.75        & 38.17      & 26.09      & 51.54         & 38.28        & 20.21      & 14.89               \\
                       & RoLT        & 81.62             & 76.58            & 42.95          & 32.59    & 60.11             & 44.23            & 23.51          & 16.61                    \\
                       & ULC         & 84.46         & 83.25        & 54.91      & 44.66   & 45.22         & 50.56        & 33.41      & 25.69                  \\
                      & MW-Net         & 70.90             & 59.85     & 32.03          & 21.71    & 46.62            & 39.33          & 19.65        & 13.72                     \\
                      & TABASCO     & 85.53         & 84.83        & 56.52      & 45.98   & 62.34         & 55.76        & 36.91      & 26.25                 \\ \midrule
Ours                   & $LR^2$        & \textbf{92.21}    & \textbf{87.61}   & \textbf{74.61} & \textbf{70.81}  & \textbf{87.53}    & \textbf{85.75}   & \textbf{72.32} & \textbf{67.24}              \\
\bottomrule
\end{tabular}
% \end{sc}
\end{small}
\end{center}
\vskip -5mm
\end{table*}
%%%%%%%%%%%%%%%%%%%%%%%%%%%%%%%%%%%%%%

\textbf{Sample size calculation for class $\bm{k}$.}
When computing the losses for experts $E_2$ and $E_3$, we uniquely weigh them based on class sample sizes. What distinguishes our approach is the use of soft labels to estimate these sizes. 
Our soft label design incorporates probabilities for each sample's class membership. To determine the sample size for class $k$, we sum the probabilities of all samples belonging to that class, fully utilizing the characteristics of soft labels and better modeling the true distribution of class sample sizes. Let the soft label ${\bm{\hat{y}}_i}=[{y_i^1}, {y_i^2}, \dots, {y_i^k}, \dots, {y_i^K}]$. The sample size of class $k$ is:
%\vspace{-3mm}
\begin{equation} 
{n_k}={\sum_{i=1}^N}\,{y_i^k}.
\vspace{-4mm}
\end{equation}

%%%%%%%%%%%%%%%%%%%%%%%%%%%%%%%%%%%%%%
\begin{table*}[t]
\caption{Test accuracy (\%) on simulated CIFAR-10 and CIFAR-100 with {\bf asymmetric} noise. Bold indicates the highest score.
\vspace{-3mm}
}
\label{table:asym}
%\vskip 0.15in
\begin{center}
\begin{small}
% \begin{sc}
\begin{tabular}{l|l|cccc|cccc}
\toprule
\multicolumn{2}{c}{Imbalance Ratio}& \multicolumn{4}{|c}{10} & \multicolumn{4}{|c}{100}\\ \midrule
\multicolumn{2}{c}{Dataset}& \multicolumn{2}{|c}{CIFAR-10} & \multicolumn{2}{c}{CIFAR-100}  & \multicolumn{2}{|c}{CIFAR-10} & \multicolumn{2}{c}{CIFAR-100}\\ \midrule
\multicolumn{2}{c|}{Noise Rate \textbf{(Asym.)}}& 0.2           & 0.4          & 0.2        & 0.4  & 0.2           & 0.4          & 0.2        & 0.4 \\ \midrule
Baseline             & CE          & 79.90         & 62.88        & 44.45      & 32.05 & 56.56    & 44.64        & 25.35      & 17.89\\ \midrule
\multirow{3}{*}{LT}    & LA          & 71.49         & 59.88        & 39.34      & 28.49 & 58.78         & 59.88        & 39.34      & 28.49\\
                       & LDAM        & 74.58         & 62.29        & 40.06      & 33.26 & 61.25         & 40.85        & 29.22      & 18.65\\
                       & IB         & 73.49      & 58.36    & 45.02          & 35.25    & 56.28   & 42.96            & 31.15          & 23.40      \\ \midrule
\multirow{2}{*}{NL}    & DivideMix   & 80.92    & 69.35        & 58.09      & 41.99   & 41.12         & 42.79    & 38.46      & 29.69 \\
                       & UNICON      & 72.81      & 69.04    & 55.99      & 44.70  & 53.53         & 34.05        & 34.14      & 30.72   \\ \midrule
\multirow{5}{*}{NL-LT} & HAR         & 82.85         & 69.19        & 48.50      & 33.20 & 62.42       & 51.97        & 27.90      & 20.03 \\
                       & RoLT        & 73.30             & 58.29            & 48.19          & 39.32   & 54.81             & 50.26            & 32.96          & -    \\
                       & ULC         & 74.07         & 73.19        & 54.45      & 43.20  & 41.14         & 22.73        & 34.07      & 25.04 \\
                      & MW-Net         & 79.34           & 65.49      & 42.52        & 30.42  & 62.19            & 45.21           & 27.56          & 20.04  \\
                      & TABASCO     & 82.10        & 80.57     & 59.39      & 50.51  & 62.98         & 54.04        & 40.35      & 33.15 \\ \midrule
Ours                   & $LR^2$        & \textbf{94.19}    & \textbf{89.67}   & \textbf{77.05} & \textbf{66.49}  & \textbf{90.92}    & \textbf{89.14}   & \textbf{73.91} & \textbf{63.26} \\
\bottomrule
\end{tabular}
% \end{sc}
\end{small}
\end{center}
\vskip -3mm
\end{table*}
%%%%%%%%%%%%%%%%%%%%%%%%%%%%%%%%%%%%%%

%%%%%%%%%%%%%%%%%%%%%%%%%%%%%%%%%%%%%%
\begin{table*}[t]
\caption{Test accuracy (\%) on Food-101N and Animal-10N with varying imbalance ratios. Bold indicates the highest score.\vspace{-2mm}
}
\label{table:food}
%\vskip 0.15in
\begin{center}
\begin{small}
% \begin{sc}
\begin{tabular}{l|l|ccc|ccc}
\toprule
\multicolumn{2}{c}{Dataset} & \multicolumn{3}{|c}{Food-101N} & \multicolumn{3}{|c}{Animal-10N}\\ \midrule
\multicolumn{2}{c|}{Imbalance Ratio}& 20      & 50          & 100   & 20      & 50          & 100\\ \midrule
Baseline             & CE          & 57.21         & 49.94        & 44.71   & 66.10         & 59.94        & 53.02 \\ \midrule
\multirow{4}{*}{LT}    & LA          & 62.81         & 55.42        & 52.30   & 69.08         & 67.78        & 61.89 \\
                       & LDAM        & 61.35         & 59.29        & 48.61  & 75.40         & 72.82        & 68.21 \\
                       & BBN         & 63.44             & 57.89            & 53.16   & 72.14             & 70.26            & 60.08 \\
                       & LWS         & 61.29         & 54.42            & 51.10 & 71.16             & 69.35            & 62.40 \\ \midrule
\multirow{3}{*}{NL}    & DivideMix   & 69.46     &57.15       &42.80   & 72.43         & 65.77        & 47.60  \\
                       & Co-learning      & 53.76   &45.92      &35.10  & 61.70         & 52.76        & 43.23   \\
                       & JoCoR      & 49.07   &32.98      &33.49   & 51.29         & 44.02        & 37.19  \\ \midrule
\multirow{4}{*}{NL-LT} & HAR         & 59.95         & 52.45        & 46.12   & 71.92         & 68.43        & 62.19   \\
                       & CL+LA           & 50.16             & 42.18            & 39.13    & 54.14           & 46.23            & 41.92   \\
                       & Co-teaching-WBL      & 58.04        & 52.12           & 53.97     & 72.43          & 71.06    & 66.60    \\
                      & H2E         & 70.35         & 63.69         & 58.66  & 77.04             & 74.94            & 66.58   \\ \midrule
Ours                   & $LR^2$        & \textbf{77.74}    & \textbf{73.14}   & \textbf{70.60}   & \textbf{81.40}    & \textbf{78.88}   & \textbf{75.48}   \\
\bottomrule
\end{tabular}
% \end{sc}
\end{small}
\end{center}
\vskip -4mm
\end{table*}
\vspace{-2mm}
\section{Experiments}

\subsection{Experimental Setup}

\textbf{Datasets.}
To comprehensively evaluate the effectiveness of the proposed method, we conducted experiments in both simulated and real-world noisy scenarios featuring long-tail distributions. Following the settings from prior long-tailed learning works~\cite{cui2019class,cao2019learning}, we simulated realistic long-tailed distributions characterized by exponential decay in sample sizes across classes. The imbalance factor, indicating the ratio between the sizes of the largest and smallest classes, was considered.
For the real-world noisy and long-tailed datasets, we utilized the Animal-10N~\cite{song2019selfie} and Food-101N~\cite{lee2018cleannet} datasets, inherently containing real-world noise.

For the simulated noisy and long-tailed datasets, we created scenarios with simulated noise on CIFAR-10~\cite{krizhevsky2009learning} and CIFAR-100~\cite{krizhevsky2009learning}. After establishing a long-tail distribution, we introduced artificially annotated noisy labels in two settings: {\em symmetric} and {\em asymmetric} noise~\cite{patrini2017making,tanaka2018joint}. Symmetric noise involved randomly replacing training data labels with all possible labels. Asymmetric noise was introduced through label flipping, such as replacing truck with automobile, bird with airplane, deer with horse, and cat with dog in CIFAR-10. Similar flipping occurs within super-classes in CIFAR-100. The noise ratio represents the ratio of noisy samples to the total sample count.

\textbf{Baselines.}
We compare our method with the following three types of approaches: (1) Long-tail learning methods (LT): LA~\cite{menon2020long}, LDAM~\cite{cao2019learning}, BBN~\cite{zhou2020bbn}, LWS~\cite{kang2019decoupling}, and IB~\cite{park2021influence}; (2) Label-noise learning methods (NL): DivideMix~\cite{li2020dividemix}, Co-learning~\cite{tan2021co}, JoCoR~\cite{wei2020combating}, and UNICON~\cite{karim2022unicon}; (3) Methods aiming at tackling noisy label and long-tail distribution (NL-LT): HAR~\cite{cao2020heteroskedastic}, RoLT~\cite{wei2021robust}, ULC~\cite{huang2022uncertainty}, MW-Net~\cite{shu2019meta}, H2E~\cite{yi2022identifying}, and TABASCO~\cite{lu2023label}. Additional comparisons with RCAL~\cite{zhang2023noisy} is provided in Appendix~\ref{appendix:rcal}.

\textbf{Implementation details.}
For the CIFAR datasets and Animal-10N, we employ ResNet18~\cite{he2016deep} as the backbone, while ResNet50~\cite{he2016deep} serves as the backbone for Food-101N. The initial learning rate is set to 0.02 for all datasets, except for the Animal-10N dataset, where it is set to 0.001. We employ the cosine annealing scheduler as the default learning rate decay strategy. Training is conducted from scratch using SGD with a weight decay of $5\times10^{-4}$ and momentum of 0.9. In stage 1, the batch size is 128, and in stage 2, it is increased to 512. Each training stage lasts for 200 epochs. For our method across all benchmarks, we configure parameters as follows: $c=6$, $\alpha=0.2$, and $\tau=0.2$. Comparative results are sourced from TABASCO~\cite{lu2023label} and H2E~\cite{yi2022identifying}. Further discussion on the influence of the scaling coefficient $c$ is provided in Appendix~\ref{appendix:c}. 

%%%%%%%%%%%%%%%%%%%%%%%%%%%%%%%%%
%%%%%%%%%%%%%%%%%%%%%%%%%%%%%%%%%
\vspace{-1mm}
\subsection{Results}

\textbf{Results on simulated CIFAR-10/100.} Table~\ref{table:sym} presents accuracy comparisons of multiple methods with various symmetric noise rates on the simulated CIFAR-10/100 datasets. Our proposed Label Refurbishment Considering Label Rarity ($LR^2$) method outperforms all baselines under an imbalance ratio of 10, across different noise rates, for both CIFAR-10 and CIFAR-100. Notably, on the more challenging CIFAR-100 dataset, our method outperforms TABASCO by a large difference of more than 10\%.
Table~\ref{table:asym} compares methods with various asymmetric noise rates of 0.2 and 0.4 under imbalance ratio 10. Our method consistently achieves superior performance in all scenarios, exhibiting a substantial gap of 20\% to 40\% compared to other methods. Moreover, as the noise rate increases, our method experiences relatively smaller performance degradation, validating its robust noise-tolerance capability within a long-tailed distribution. This suggests that methods focusing on identifying and segregating noisy data may witness reduced ability to recognize noisy samples as the imbalance ratio increases, leading to potential errors in sample handling and consequent performance degradation.

\textbf{Results on real-world noisy and imbalanced datasets.} 
Table~\ref{table:food} shows evaluation on Food-101N and Animal-10N. Across all scenarios, our method consistently outperforms other state-of-the-art methods. Particularly on the Food-101N dataset with larger-scale and higher noise rate, our approach demonstrates the most robust performance, surpassing other methods as the imbalance ratio increases. These results affirm the efficacy of our proposed approach in effectively addressing the challenges posed by real-world noise and unbalanced datasets.

%%%%%%%%%%%%%%%%%%%%%%%%%%%%%
%%%%%%%%%%%%%%%%%%%%%%%%%%%%%
\vspace{-0mm}
\subsection{Ablation Study and Discussions}

\textbf{Ablation studies on simulated CIFAR-100.}
We conducted a comprehensive ablation study on the simulated CIFAR-100 dataset with an asymmetric noise rate of 0.4, analyzing each component of our method under imbalance ratios of 10 and 100. Table~\ref{table:ablation_all} reveals the following insights:
(1) Regarding contrastive learning baseline, under noisy labels and a long-tailed distribution, self-supervised contrastive learning demonstrates robust performance.
(2) Our proposed BANC loss, built upon contrastive learning, further improves performance by 6\%, providing enhanced initial predictions.
(3) The application of our label refurbishment and multi-expert ensemble learning leads to an additional improvement of around 3\%. Even when using the less effective cross-entropy loss in the first stage, label refurbishment and ensemble learning result in a 5\% improvement, highlighting the effectiveness and stability of these components.
(4) Notably, label refurbishment itself alone plays a crucial role, as evident when performing ensemble learning without it; the effect decreases significantly by 21\% and 28\% under imbalance ratios of 10 and 100, respectively. This result emphasizes the importance of label refurbishment in addressing incorrect labels, which, if not corrected, can greatly impact class sample size statistics and subsequently influence weighted loss during expert training.

%We conducted a detailed ablation study on the simulated CIFAR100 dataset with an asymmetric noise rate of 0.4 to analyze each component of our method. The experiments were performed under imbalance ratios of 10 and 100, respectively. As shown in Table \ref{table:ablation_all}, first, we observe that under noisy labels and long-tailed distribution, self-supervised contrastive learning demonstrated robust performance. Based on contrastive learning, our proposed BANC loss improved performance by 6\%, providing better initial predictions. Next, when applying our proposed label refurbishment strategy and training with multi-expert ensemble learning, performance improved further by around 3\%. If we use the less effective cross-entropy loss in the first stage, label refurbishment and ensemble learning also cause an improvement 5\% in the model, demonstrating the effectiveness and stability of these two components. In particular, label refinement plays a crucial role, as seen when performing ensemble learning without label refurbishment; the effect significantly decreased by 21\% and 28\% under the imbalance ratio of 10 and 100, respectively. This is because the loss of training experts is weighted by the class sample size, and without label refurbishment, incorrect labels persist, affecting the statistics of class sample size and subsequently influencing the weighted loss.

%%%%%%%%%%%%%%%%%%%%%%%%%%%%%%
%%%%%%%%%%%%%%%%%%%%%%%%%%%%%%%%%%%%%%
\begin{table}[t]
\caption{Ablation study of test accuracy (\%) on simulated CIFAR-100. CL: self-supervised contrastive learning. BANC: BAlanced Noise-tolerant Cross entropy, Re-Label: label refurbishment, Multi-Exp: multi-expert ensemble Learning. Symbol \XSolidBrush in CL denotes training with standard supervised learning. Symbol \XSolidBrush in BANC indicates that cross-entropy loss is used.
\vspace{-4mm}
}
\label{table:ablation_all}
%\vskip 0.15in
\begin{center}
\begin{small}
% \begin{sc}
\begin{tabular}{cccc|cc}
\toprule
\multicolumn{4}{c}{Component} &\multicolumn{2}{|c}{Imbalance Ratio} \\ \midrule
CL &BANC     &Re-Label   &Multi-Expert  & 10 & 100   \\ \midrule
\XSolidBrush &\XSolidBrush   &\XSolidBrush    &\XSolidBrush    &15.51   &17.89   \\
\Checkmark &\XSolidBrush   &\XSolidBrush    &\XSolidBrush    &57.43   &55.98   \\ 
\Checkmark  &\XSolidBrush   &\Checkmark    &\Checkmark    &64.29   &60.91  \\
\Checkmark &\Checkmark   &\XSolidBrush    &\XSolidBrush    &63.30   &59.26  \\
\Checkmark &\Checkmark   &\XSolidBrush    &\Checkmark    &41.95   &30.87   \\
\Checkmark &\Checkmark   &\Checkmark    &\Checkmark    &\textbf{66.49}   &\textbf{63.26}   \\
\bottomrule
\end{tabular}
% \end{sc}
\end{small}
\end{center}
\vskip -3mm
\end{table}
%%%%%%%%%%%%%%%%%%%%%%%%%%%%%%%%%%%%%%
%%%%%%%%%%%%%%%%%%%%%%%%%%%%%%%%%%%%%%
\begin{table}[t]
\caption{Test accuracy (\%) on the simulated CIFAR-100 with an imbalance ratio of 100. The noise rates are asymmetric at 0.4 and symmetric at 0.6, respectively.
\vspace{-2mm}
}
\label{table:relabel}
%\vskip 0.15in
\begin{center}
\begin{small}
% \begin{sc}
\begin{tabular}{l|cc}
\toprule
Stage    &\textbf{Asym.} 0.4    &\textbf{Sym.} 0.6      \\    \midrule
Stage 1                          &55.98     &66.63      \\
Stage 2 w/o re-label    &57.15    &67.48      \\ 
Stage 2                          &\textbf{62.24}    &\textbf{67.76}     \\
\bottomrule
\end{tabular}
% \end{sc}
\end{small}
\end{center}
\vskip -3mm
\end{table}
%%%%%%%%%%%%%%%%%%%%%%%%%%%%%%%%%%%%%%
%%%%%%%%%%%%%%%%%%%%%%%%%%%%%%%%%

%The following ablation studies are conducted to further investigate the effectiveness of the label refurbishment and multi-expert ensemble learning, individually.

\textbf{Effectiveness of the label refurbishment strategy.}
Additional experiments are conducted on simulated CIFAR-100 with imbalance ratio of 100 to assess label refurbishment. We compared results from the first-stage predictions, second-stage predictions with/without label refurbishment. As discussed earlier, not applying label refurbishment impacts the loss weighted by class sample size. In the second stage, we employed a classifier trained solely with cross-entropy for a more precise evaluation. The results, presented in Table~\ref{table:relabel}, reveal an improvement of approximately 2\% in the performance of the second-stage predictions with label refurbishment compared to those without label refurbishment, affirming the effectiveness of the proposed strategy.
%%%%%%%%%%%%%%%%%%%%%%%%%%%%%%%%%%%%%%
\begin{table}[t]
\caption{Test accuracy (\%) of many/medium/few classes on the simulated CIFAR-100 with an asymmetric noise rate of 0.4 and imbalance ratio of 100.
\vspace{-2mm}
}
\label{table:expertasym}
%\vskip 0.15in
\begin{center}
\begin{small}
% \begin{sc}
\begin{tabular}{c|cccc}
\toprule
Model &Many     &Medium   &Few  & All   \\ \midrule
Expert $E_1$    &\textbf{92.82}    &66.27    &8.14   &59.31   \\ 
Expert $E_2$      &84.10    &67.46    &18.90   &61.15  \\
Expert $E_3$     &57.95    &48.81    &\textbf{21.22}   &45.19  \\   \midrule
Ensemble    &81.03    &\textbf{70.48}    &19.77   &\textbf{63.26}   \\
\bottomrule
\end{tabular}
% \end{sc}
\end{small}
\end{center}
\vskip -6mm
\end{table}
%%%%%%%%%%%%%%%%%%%%%%%%%%%%%%%%%%%%%%

\textbf{Effectiveness of multi-expert ensemble learning.}
Additional experiments are conducted on simulated CIFAR-100 with imbalance ratio of 100 to validate the characteristics of the experts regarding multi-expert ensemble learning across three class subgroups: (1) many-shot classes over 100 training images, (2) medium-shot classes with 20 to 100 images, and (3) few-shot classes with less than 20 images. As shown in Table~\ref{table:expertasym}, Experts $E_1$ and $E_3$ perform exceptionally well in the many-shot and few-shot classes, respectively, while Expert $E_2$ shows average performance across all three class subgroups. Through the ensemble of the three experts, overall performance is enhanced, with notable improvements observed in all three subgroups.

%%%%%%%%%%%%%%%%%%%%%%%%%%%%%%%%%%%%%%
% \begin{table}[t]
% \caption{Test accuracy (\%) of many/medium/few classes on the simulated CIFAR-100 with an asymmetric noise rate of 0.4 and imbalance ratio of 100.
% \vspace{-2mm}
% }
% \label{table:expertasym}
% %\vskip 0.15in
% \begin{center}
% \begin{small}
% % \begin{sc}
% \begin{tabular}{c|cccc}
% \toprule
% Model &Many     &Medium   &Few  & All   \\ \midrule
% Expert $E_1$    &\textbf{92.82}    &66.27    &8.14   &59.31   \\ 
% Expert $E_2$      &84.10    &67.46    &18.90   &61.15  \\
% Expert $E_3$     &57.95    &48.81    &\textbf{21.22}   &45.19  \\   \midrule
% Ensemble    &81.03    &\textbf{70.48}    &19.77   &\textbf{63.26}   \\
% \bottomrule
% \end{tabular}
% % \end{sc}
% \end{small}
% \end{center}
% \vskip -6mm
% \end{table}
%%%%%%%%%%%%%%%%%%%%%%%%%%%%%%%%%%%%%%
%%%%%%%%%%%%%%%%%%%%%%%%%%%%%%%%%

\section{Conclusion}

In this paper, we introduced a two-stage approach to address the challenges of long-tailed noisy label classification. Our innovative strategy combines a soft label refurbishment technique with multi-expert ensemble learning. Noisy labels are managed through soft-label refurbishment, utilizing a noise-tolerant loss function. The long-tailed issue is tackled using three expert classifiers, each specializing in one of the categories (many-shot, medium-shot, and few-shot sub-groups). Adaptive loss functions are designed for each category, with weights tailored to match the frequency. Compared to the prior bi-dimensional sample selection and representation calibration method, our approach exhibits improved robustness and generalizability. Extensive experimental results validate the effectiveness of our two-stage approach.

%%%%%%%%%%%%%%%%%%%%%%%%%%%%%%%%%%%%%%
%%%%%%%%%%%%%%%%%%%%%%%%%%%%%%%%%%%%%%
% In the unusual situation where you want a paper to appear in the
% references without citing it in the main text, use \nocite
\nocite{langley00}

\bibliography{paper}

\begin{thebibliography}{58}
\providecommand{\natexlab}[1]{#1}
\providecommand{\url}[1]{\texttt{#1}}
\expandafter\ifx\csname urlstyle\endcsname\relax
  \providecommand{\doi}[1]{doi: #1}\else
  \providecommand{\doi}{doi: \begingroup \urlstyle{rm}\Url}\fi

\bibitem[Byrd \& Lipton(2019)Byrd and Lipton]{byrd2019effect}
Byrd, J. and Lipton, Z.
\newblock What is the effect of importance weighting in deep learning?
\newblock In \emph{International conference on machine learning}, pp.\  872--881. PMLR, 2019.

\bibitem[Cao et~al.(2019)Cao, Wei, Gaidon, Arechiga, and Ma]{cao2019learning}
Cao, K., Wei, C., Gaidon, A., Arechiga, N., and Ma, T.
\newblock Learning imbalanced datasets with label-distribution-aware margin loss.
\newblock \emph{Advances in neural information processing systems}, 32, 2019.

\bibitem[Cao et~al.(2020)Cao, Chen, Lu, Arechiga, Gaidon, and Ma]{cao2020heteroskedastic}
Cao, K., Chen, Y., Lu, J., Arechiga, N., Gaidon, A., and Ma, T.
\newblock Heteroskedastic and imbalanced deep learning with adaptive regularization.
\newblock \emph{arXiv preprint arXiv:2006.15766}, 2020.

\bibitem[Chen et~al.(2023)Chen, Cheng, Du, Xu, Jiang, and Wang]{chen2023two}
Chen, M., Cheng, H., Du, Y., Xu, M., Jiang, W., and Wang, C.
\newblock Two wrongs don’t make a right: Combating confirmation bias in learning with label noise.
\newblock In \emph{Proceedings of the AAAI Conference on Artificial Intelligence}, volume~37, pp.\  14765--14773, 2023.

\bibitem[Chen et~al.(2021)Chen, Ye, Chen, Zhao, and Heng]{chen2021beyond}
Chen, P., Ye, J., Chen, G., Zhao, J., and Heng, P.-A.
\newblock Beyond class-conditional assumption: A primary attempt to combat instance-dependent label noise.
\newblock In \emph{Proceedings of the AAAI Conference on Artificial Intelligence}, volume~35, pp.\  11442--11450, 2021.

\bibitem[Chen \& He(2021)Chen and He]{chen2021exploring}
Chen, X. and He, K.
\newblock Exploring simple siamese representation learning.
\newblock In \emph{Proceedings of the IEEE/CVF conference on computer vision and pattern recognition}, pp.\  15750--15758, 2021.

\bibitem[Chen et~al.(2020)Chen, Fan, Girshick, and He]{chen2020improved}
Chen, X., Fan, H., Girshick, R., and He, K.
\newblock Improved baselines with momentum contrastive learning.
\newblock \emph{arXiv preprint arXiv:2003.04297}, 2020.

\bibitem[Chou et~al.(2020)Chou, Chang, Pan, Wei, and Juan]{chou2020remix}
Chou, H.-P., Chang, S.-C., Pan, J.-Y., Wei, W., and Juan, D.-C.
\newblock Remix: rebalanced mixup.
\newblock In \emph{Computer Vision--ECCV 2020 Workshops: Glasgow, UK, August 23--28, 2020, Proceedings, Part VI 16}, pp.\  95--110. Springer, 2020.

\bibitem[Cui et~al.(2019)Cui, Jia, Lin, Song, and Belongie]{cui2019class}
Cui, Y., Jia, M., Lin, T.-Y., Song, Y., and Belongie, S.
\newblock Class-balanced loss based on effective number of samples.
\newblock In \emph{Proceedings of the IEEE/CVF conference on computer vision and pattern recognition}, pp.\  9268--9277, 2019.

\bibitem[Devlin et~al.(2018)Devlin, Chang, Lee, and Toutanova]{devlin2018bert}
Devlin, J., Chang, M.-W., Lee, K., and Toutanova, K.
\newblock Bert: Pre-training of deep bidirectional transformers for language understanding.
\newblock \emph{arXiv preprint arXiv:1810.04805}, 2018.

\bibitem[Du \& Wu(2023)Du and Wu]{du2023no}
Du, Y. and Wu, J.
\newblock No one left behind: Improving the worst categories in long-tailed learning.
\newblock In \emph{Proceedings of the IEEE/CVF Conference on Computer Vision and Pattern Recognition}, pp.\  15804--15813, 2023.

\bibitem[Estabrooks et~al.(2004)Estabrooks, Jo, and Japkowicz]{estabrooks2004multiple}
Estabrooks, A., Jo, T., and Japkowicz, N.
\newblock A multiple resampling method for learning from imbalanced data sets.
\newblock \emph{Computational intelligence}, 20\penalty0 (1):\penalty0 18--36, 2004.

\bibitem[Ghosh \& Lan(2021)Ghosh and Lan]{ghosh2021contrastive}
Ghosh, A. and Lan, A.
\newblock Contrastive learning improves model robustness under label noise.
\newblock In \emph{Proceedings of the IEEE/CVF Conference on Computer Vision and Pattern Recognition}, pp.\  2703--2708, 2021.

\bibitem[Haixiang et~al.(2017)Haixiang, Yijing, Shang, Mingyun, Yuanyue, and Bing]{haixiang2017learning}
Haixiang, G., Yijing, L., Shang, J., Mingyun, G., Yuanyue, H., and Bing, G.
\newblock Learning from class-imbalanced data: Review of methods and applications.
\newblock \emph{Expert systems with applications}, 73:\penalty0 220--239, 2017.

\bibitem[Han et~al.(2018)Han, Yao, Yu, Niu, Xu, Hu, Tsang, and Sugiyama]{han2018co}
Han, B., Yao, Q., Yu, X., Niu, G., Xu, M., Hu, W., Tsang, I., and Sugiyama, M.
\newblock Co-teaching: Robust training of deep neural networks with extremely noisy labels.
\newblock \emph{Advances in neural information processing systems}, 31, 2018.

\bibitem[He et~al.(2016)He, Zhang, Ren, and Sun]{he2016deep}
He, K., Zhang, X., Ren, S., and Sun, J.
\newblock Deep residual learning for image recognition.
\newblock In \emph{Proceedings of the IEEE conference on computer vision and pattern recognition}, pp.\  770--778, 2016.

\bibitem[He et~al.(2021)He, Wu, and Wei]{he2021distilling}
He, Y.-Y., Wu, J., and Wei, X.-S.
\newblock Distilling virtual examples for long-tailed recognition.
\newblock In \emph{Proceedings of the IEEE/CVF International Conference on Computer Vision}, pp.\  235--244, 2021.

\bibitem[Hendrycks et~al.(2018)Hendrycks, Mazeika, Wilson, and Gimpel]{hendrycks2018using}
Hendrycks, D., Mazeika, M., Wilson, D., and Gimpel, K.
\newblock Using trusted data to train deep networks on labels corrupted by severe noise.
\newblock \emph{Advances in neural information processing systems}, 31, 2018.

\bibitem[Hong et~al.(2021)Hong, Han, Choi, Seo, Kim, and Chang]{hong2021disentangling}
Hong, Y., Han, S., Choi, K., Seo, S., Kim, B., and Chang, B.
\newblock Disentangling label distribution for long-tailed visual recognition.
\newblock In \emph{Proceedings of the IEEE/CVF conference on computer vision and pattern recognition}, pp.\  6626--6636, 2021.

\bibitem[Huang et~al.(2022)Huang, Bai, Zhao, Bai, and Wang]{huang2022uncertainty}
Huang, Y., Bai, B., Zhao, S., Bai, K., and Wang, F.
\newblock Uncertainty-aware learning against label noise on imbalanced datasets.
\newblock In \emph{Proceedings of the AAAI Conference on Artificial Intelligence}, volume~36, pp.\  6960--6969, 2022.

\bibitem[Kang et~al.(2019)Kang, Xie, Rohrbach, Yan, Gordo, Feng, and Kalantidis]{kang2019decoupling}
Kang, B., Xie, S., Rohrbach, M., Yan, Z., Gordo, A., Feng, J., and Kalantidis, Y.
\newblock Decoupling representation and classifier for long-tailed recognition.
\newblock \emph{arXiv preprint arXiv:1910.09217}, 2019.

\bibitem[Karim et~al.(2022)Karim, Rizve, Rahnavard, Mian, and Shah]{karim2022unicon}
Karim, N., Rizve, M.~N., Rahnavard, N., Mian, A., and Shah, M.
\newblock Unicon: Combating label noise through uniform selection and contrastive learning.
\newblock In \emph{Proceedings of the IEEE/CVF Conference on Computer Vision and Pattern Recognition}, pp.\  9676--9686, 2022.

\bibitem[Krizhevsky et~al.(2009)Krizhevsky, Hinton, et~al.]{krizhevsky2009learning}
Krizhevsky, A., Hinton, G., et~al.
\newblock Learning multiple layers of features from tiny images.
\newblock 2009.

\bibitem[Krizhevsky et~al.(2012)Krizhevsky, Sutskever, and Hinton]{krizhevsky2012imagenet}
Krizhevsky, A., Sutskever, I., and Hinton, G.~E.
\newblock Imagenet classification with deep convolutional neural networks.
\newblock \emph{Advances in neural information processing systems}, 25, 2012.

\bibitem[Lee et~al.(2021)Lee, Kim, Lee, Lee, and Choo]{lee2021learning}
Lee, J., Kim, E., Lee, J., Lee, J., and Choo, J.
\newblock Learning debiased representation via disentangled feature augmentation.
\newblock \emph{Advances in Neural Information Processing Systems}, 34:\penalty0 25123--25133, 2021.

\bibitem[Lee et~al.(2018)Lee, He, Zhang, and Yang]{lee2018cleannet}
Lee, K.-H., He, X., Zhang, L., and Yang, L.
\newblock Cleannet: Transfer learning for scalable image classifier training with label noise.
\newblock In \emph{Proceedings of the IEEE conference on computer vision and pattern recognition}, pp.\  5447--5456, 2018.

\bibitem[Li et~al.(2019)Li, Wong, Zhao, and Kankanhalli]{li2019learning}
Li, J., Wong, Y., Zhao, Q., and Kankanhalli, M.~S.
\newblock Learning to learn from noisy labeled data.
\newblock In \emph{Proceedings of the IEEE/CVF conference on computer vision and pattern recognition}, pp.\  5051--5059, 2019.

\bibitem[Li et~al.(2020)Li, Socher, and Hoi]{li2020dividemix}
Li, J., Socher, R., and Hoi, S.~C.
\newblock Dividemix: Learning with noisy labels as semi-supervised learning.
\newblock In \emph{ICLR 2020}, 2020.

\bibitem[Liu et~al.(2021)Liu, HaoChen, Gaidon, and Ma]{liu2021self}
Liu, H., HaoChen, J.~Z., Gaidon, A., and Ma, T.
\newblock Self-supervised learning is more robust to dataset imbalance.
\newblock \emph{arXiv preprint arXiv:2110.05025}, 2021.

\bibitem[Lu et~al.(2023)Lu, Zhang, Han, Cheung, and Wang]{lu2023label}
Lu, Y., Zhang, Y., Han, B., Cheung, Y.-m., and Wang, H.
\newblock Label-noise learning with intrinsically long-tailed data.
\newblock In \emph{Proceedings of the IEEE/CVF International Conference on Computer Vision}, pp.\  1369--1378, 2023.

\bibitem[Ma et~al.(2018)Ma, Wang, Houle, Zhou, Erfani, Xia, Wijewickrema, and Bailey]{ma2018dimensionality}
Ma, X., Wang, Y., Houle, M.~E., Zhou, S., Erfani, S., Xia, S., Wijewickrema, S., and Bailey, J.
\newblock Dimensionality-driven learning with noisy labels.
\newblock In \emph{International Conference on Machine Learning}, pp.\  3355--3364. PMLR, 2018.

\bibitem[Menon et~al.(2020)Menon, Jayasumana, Rawat, Jain, Veit, and Kumar]{menon2020long}
Menon, A.~K., Jayasumana, S., Rawat, A.~S., Jain, H., Veit, A., and Kumar, S.
\newblock Long-tail learning via logit adjustment.
\newblock \emph{arXiv preprint arXiv:2007.07314}, 2020.

\bibitem[Park et~al.(2021)Park, Lim, Jeon, and Choi]{park2021influence}
Park, S., Lim, J., Jeon, Y., and Choi, J.~Y.
\newblock Influence-balanced loss for imbalanced visual classification.
\newblock In \emph{Proceedings of the IEEE/CVF International Conference on Computer Vision}, pp.\  735--744, 2021.

\bibitem[Patrini et~al.(2017)Patrini, Rozza, Krishna~Menon, Nock, and Qu]{patrini2017making}
Patrini, G., Rozza, A., Krishna~Menon, A., Nock, R., and Qu, L.
\newblock Making deep neural networks robust to label noise: A loss correction approach.
\newblock In \emph{Proceedings of the IEEE conference on computer vision and pattern recognition}, pp.\  1944--1952, 2017.

\bibitem[Pereyra et~al.(2017)Pereyra, Tucker, Chorowski, Kaiser, and Hinton]{pereyra2017regularizing}
Pereyra, G., Tucker, G., Chorowski, J., Kaiser, {\L}., and Hinton, G.
\newblock Regularizing neural networks by penalizing confident output distributions.
\newblock \emph{arXiv preprint arXiv:1701.06548}, 2017.

\bibitem[Ren et~al.(2020)Ren, Yu, Ma, Zhao, Yi, et~al.]{ren2020balanced}
Ren, J., Yu, C., Ma, X., Zhao, H., Yi, S., et~al.
\newblock Balanced meta-softmax for long-tailed visual recognition.
\newblock \emph{Advances in neural information processing systems}, 33:\penalty0 4175--4186, 2020.

\bibitem[Ren et~al.(2015)Ren, He, Girshick, and Sun]{ren2015faster}
Ren, S., He, K., Girshick, R., and Sun, J.
\newblock Faster r-cnn: Towards real-time object detection with region proposal networks.
\newblock \emph{Advances in neural information processing systems}, 28, 2015.

\bibitem[Shu et~al.(2019)Shu, Xie, Yi, Zhao, Zhou, Xu, and Meng]{shu2019meta}
Shu, J., Xie, Q., Yi, L., Zhao, Q., Zhou, S., Xu, Z., and Meng, D.
\newblock Meta-weight-net: Learning an explicit mapping for sample weighting.
\newblock \emph{Advances in neural information processing systems}, 32, 2019.

\bibitem[Song et~al.(2019)Song, Kim, and Lee]{song2019selfie}
Song, H., Kim, M., and Lee, J.-G.
\newblock Selfie: Refurbishing unclean samples for robust deep learning.
\newblock In \emph{International Conference on Machine Learning}, pp.\  5907--5915. PMLR, 2019.

\bibitem[Song et~al.(2022)Song, Kim, Park, Shin, and Lee]{song2022learning}
Song, H., Kim, M., Park, D., Shin, Y., and Lee, J.-G.
\newblock Learning from noisy labels with deep neural networks: A survey.
\newblock \emph{IEEE Transactions on Neural Networks and Learning Systems}, 2022.

\bibitem[Tan et~al.(2021)Tan, Xia, Wu, and Li]{tan2021co}
Tan, C., Xia, J., Wu, L., and Li, S.~Z.
\newblock Co-learning: Learning from noisy labels with self-supervision.
\newblock In \emph{Proceedings of the 29th ACM International Conference on Multimedia}, pp.\  1405--1413, 2021.

\bibitem[Tanaka et~al.(2018)Tanaka, Ikami, Yamasaki, and Aizawa]{tanaka2018joint}
Tanaka, D., Ikami, D., Yamasaki, T., and Aizawa, K.
\newblock Joint optimization framework for learning with noisy labels.
\newblock In \emph{Proceedings of the IEEE conference on computer vision and pattern recognition}, pp.\  5552--5560, 2018.

\bibitem[Tao et~al.(2023)Tao, Sun, Yang, Chen, Wang, Yang, Du, and Zheng]{tao2023local}
Tao, Y., Sun, J., Yang, H., Chen, L., Wang, X., Yang, W., Du, D., and Zheng, M.
\newblock Local and global logit adjustments for long-tailed learning.
\newblock In \emph{Proceedings of the IEEE/CVF International Conference on Computer Vision}, pp.\  11783--11792, 2023.

\bibitem[Wang et~al.(2020)Wang, Lian, Miao, Liu, and Yu]{wang2020long}
Wang, X., Lian, L., Miao, Z., Liu, Z., and Yu, S.~X.
\newblock Long-tailed recognition by routing diverse distribution-aware experts.
\newblock \emph{arXiv preprint arXiv:2010.01809}, 2020.

\bibitem[Wang et~al.(2019)Wang, Ma, Chen, Luo, Yi, and Bailey]{wang2019symmetric}
Wang, Y., Ma, X., Chen, Z., Luo, Y., Yi, J., and Bailey, J.
\newblock Symmetric cross entropy for robust learning with noisy labels.
\newblock In \emph{Proceedings of the IEEE/CVF international conference on computer vision}, pp.\  322--330, 2019.

\bibitem[Wei et~al.(2020)Wei, Feng, Chen, and An]{wei2020combating}
Wei, H., Feng, L., Chen, X., and An, B.
\newblock Combating noisy labels by agreement: A joint training method with co-regularization.
\newblock In \emph{Proceedings of the IEEE/CVF conference on computer vision and pattern recognition}, pp.\  13726--13735, 2020.

\bibitem[Wei et~al.(2021)Wei, Shi, Tu, and Li]{wei2021robust}
Wei, T., Shi, J.-X., Tu, W.-W., and Li, Y.-F.
\newblock Robust long-tailed learning under label noise.
\newblock \emph{arXiv preprint arXiv:2108.11569}, 2021.

\bibitem[Xia et~al.(2020)Xia, Liu, Han, Gong, Wang, Ge, and Chang]{xia2020robust}
Xia, X., Liu, T., Han, B., Gong, C., Wang, N., Ge, Z., and Chang, Y.
\newblock Robust early-learning: Hindering the memorization of noisy labels.
\newblock In \emph{International conference on learning representations}, 2020.

\bibitem[Xia et~al.(2021)Xia, Liu, Han, Gong, Yu, Niu, and Sugiyama]{xia2021sample}
Xia, X., Liu, T., Han, B., Gong, M., Yu, J., Niu, G., and Sugiyama, M.
\newblock Sample selection with uncertainty of losses for learning with noisy labels.
\newblock \emph{arXiv preprint arXiv:2106.00445}, 2021.

\bibitem[Xiao et~al.(2015)Xiao, Xia, Yang, Huang, and Wang]{xiao2015learning}
Xiao, T., Xia, T., Yang, Y., Huang, C., and Wang, X.
\newblock Learning from massive noisy labeled data for image classification.
\newblock In \emph{Proceedings of the IEEE conference on computer vision and pattern recognition}, pp.\  2691--2699, 2015.

\bibitem[Yao et~al.(2021)Yao, Sun, Zhang, Shen, Wu, Zhang, and Tang]{yao2021jo}
Yao, Y., Sun, Z., Zhang, C., Shen, F., Wu, Q., Zhang, J., and Tang, Z.
\newblock Jo-src: A contrastive approach for combating noisy labels.
\newblock In \emph{Proceedings of the IEEE/CVF conference on computer vision and pattern recognition}, pp.\  5192--5201, 2021.

\bibitem[Yi et~al.(2022)Yi, Tang, Hua, Lim, and Zhang]{yi2022identifying}
Yi, X., Tang, K., Hua, X.-S., Lim, J.-H., and Zhang, H.
\newblock Identifying hard noise in long-tailed sample distribution.
\newblock In \emph{European Conference on Computer Vision}, pp.\  739--756. Springer, 2022.

\bibitem[Zang et~al.(2021)Zang, Huang, and Loy]{zang2021fasa}
Zang, Y., Huang, C., and Loy, C.~C.
\newblock Fasa: Feature augmentation and sampling adaptation for long-tailed instance segmentation.
\newblock In \emph{Proceedings of the IEEE/CVF International Conference on Computer Vision}, pp.\  3457--3466, 2021.

\bibitem[Zhang et~al.(2023{\natexlab{a}})Zhang, Zhao, Yao, Yuan, and Huang]{zhang2023noisy}
Zhang, M., Zhao, X., Yao, J., Yuan, C., and Huang, W.
\newblock When noisy labels meet long tail dilemmas: A representation calibration method.
\newblock In \emph{Proceedings of the IEEE/CVF International Conference on Computer Vision}, pp.\  15890--15900, 2023{\natexlab{a}}.

\bibitem[Zhang et~al.(2023{\natexlab{b}})Zhang, Kang, Hooi, Yan, and Feng]{zhang2023deep}
Zhang, Y., Kang, B., Hooi, B., Yan, S., and Feng, J.
\newblock Deep long-tailed learning: A survey.
\newblock \emph{IEEE Transactions on Pattern Analysis and Machine Intelligence}, 2023{\natexlab{b}}.

\bibitem[Zhao et~al.(2023)Zhao, Jiang, Hu, Zhang, and Liu]{zhao2023mdcs}
Zhao, Q., Jiang, C., Hu, W., Zhang, F., and Liu, J.
\newblock Mdcs: More diverse experts with consistency self-distillation for long-tailed recognition.
\newblock In \emph{Proceedings of the IEEE/CVF International Conference on Computer Vision}, pp.\  11597--11608, 2023.

\bibitem[Zhong et~al.(2021)Zhong, Cui, Liu, and Jia]{zhong2021improving}
Zhong, Z., Cui, J., Liu, S., and Jia, J.
\newblock Improving calibration for long-tailed recognition.
\newblock In \emph{Proceedings of the IEEE/CVF conference on computer vision and pattern recognition}, pp.\  16489--16498, 2021.

\bibitem[Zhou et~al.(2020)Zhou, Cui, Wei, and Chen]{zhou2020bbn}
Zhou, B., Cui, Q., Wei, X.-S., and Chen, Z.-M.
\newblock Bbn: Bilateral-branch network with cumulative learning for long-tailed visual recognition.
\newblock In \emph{Proceedings of the IEEE/CVF conference on computer vision and pattern recognition}, pp.\  9719--9728, 2020.

\end{thebibliography}
\bibliographystyle{icml2024}

%%%%%%%%%%%%%%%%%%%%%%%%%%%%%%%%%%%%%%%%%%%%%%%%%%%%%%%%%%%%%%%%%%%%%%%%%%%%%%%
%%%%%%%%%%%%%%%%%%%%%%%%%%%%%%%%%%%%%%%%%%%%%%%%%%%%%%%%%%%%%%%%%%%%%%%%%%%%%%%
% APPENDIX
%%%%%%%%%%%%%%%%%%%%%%%%%%%%%%%%%%%%%%%%%%%%%%%%%%%%%%%%%%%%%%%%%%%%%%%%%%%%%%%
%%%%%%%%%%%%%%%%%%%%%%%%%%%%%%%%%%%%%%%%%%%%%%%%%%%%%%%%%%%%%%%%%%%%%%%%%%%%%%%
\newpage
\appendix

\renewcommand{\theequation}{A\arabic{equation}}
\renewcommand{\thefigure}{A\arabic{figure}}
\renewcommand{\thetable}{A\arabic{table}}
\setcounter{equation}{0}    
\setcounter{figure}{0}    
\setcounter{table}{0}    

\onecolumn
\section{Comparison with Representation CALibration (RCAL)}
\label{appendix:rcal}

In the context of addressing noise labels and long-tailed distribution issues simultaneously, RCAL~\cite{zhang2023noisy}
stands out as the only method among current state-of-the-art approaches that doesn't adopt the strategy of segregating noisy and clean samples. 
Therefore, we discuss RCAL as a distinct case. Similar to our approach, RCAL initially employs contrastive learning to enhance feature learning. However, it diverges by incorporating representation calibration as a subsequent optimization step, whereas our method involves training the classifier using the proposed BAlanced Noise-tolerant Cross entropy (BANC) loss. Additionally, we combine it with a label refurbishment strategy and multi-expert ensemble learning to further address the challenges of noisy labels and long-tailed distribution.

Following the methodology outlined in the RCAL original paper, we simulated various noise rates and imbalance ratios on CIFAR-10 and CIFAR-100 datasets, conducting experiments and comparing results with those reported in the RCAL original paper. As shown in Table~\ref{table:rcal}, despite a similar feature learning approach, our method consistently outperforms RCAL in all scenarios. Notably, at imbalance ratio 100, the performance gap between the two methods is more significant, with differences of approximately 15\% and 30\% on CIFAR-10 and CIFAR-100, respectively. This highlights the effectiveness of our two-stage approach in addressing challenges related to simultaneous noisy labels and long-tailed distribution.

%%%%%%%%%%%%%%%%%%%%%%%%%%%%%%%%%%%%%%
%%%%%%%%%%%%%%%%%%%%%%%%%%%%%%%%%%%%%%
\makeatletter\def\@captype{table}\makeatother
% \begin{table}[t]
\caption{Test accuracy (\%) on simulated CIFAR-10 and CIFAR-100 with varying noise rates and imbalance ratios. Bold shows the highest score.}
\label{table:rcal}
\vskip -0.15in
\begin{center}
\begin{small}
% \begin{sc}
\begin{tabular}{c|c|ccccc|ccccc}
\toprule
Dataset    & Imbalance Ratio  &\multicolumn{5}{c}{10} &\multicolumn{5}{c}{100} \\ \midrule
\multirow{16}*{\rotatebox{90}{CIFAR-10}}    &Noise Rate    & 0.1 &0.2 & 0.3 &0.4 &0.5 & 0.1 &0.2 & 0.3 &0.4 &0.5   \\ \cmidrule{2-12}
    & ERM &80.41 &75.61 &71.94 &70.13 &63.25 &64.41 &62.17 &52.94 &48.11 &38.71  \\
    & ERM-DRW &81.72 &77.61 &71.94 &70.13 &63.25 &66.74 &62.17 &52.94 &48.11 &38.71  \\ \cmidrule{2-12}
    & LDAM &84.59 &82.37 &77.48 &71.41 &60.30 &71.46 &66.26 &58.34 &46.64 &36.66    \\
    & LDAM-DRW &85.94 &83.73 &80.20 &74.87 &67.93 &76.58 &72.28 &66.68 &57.51 &43.23 \\ 
    & CRT &80.22 &76.15 &74.17 &70.05 &64.15 &61.54 &59.52 &54.05 &50.12 &36.73 \\
    & NCM &82.33 &74.73 &74.76 &68.43 &64.82 &68.09 &66.25 &60.91 &55.47 &42.61     \\
    & MiSLAS &87.58 &85.21 &83.39 &76.16 &72.46 &75.62 &71.48 &67.90 &62.04 &54.54 \\ \cmidrule{2-12}
    & Co-teaching &80.30 &78.54 &68.71 &57.10 &46.77 &55.58 &50.29 &38.01 &30.75 &22.85 \\
    & CDR &81.68 &78.09 &73.86 &68.12 &62.24 &60.47 &55.34 &46.32 &42.51 &32.44 \\
    & Sel-CL+ &86.47 &85.11 &84.41 &80.35 &77.27 &72.31 &71.02 &65.70 &61.37 &56.21  \\ \cmidrule{2-12}
    & HAR-DRW &84.09 &82.43 &80.41 &77.43 &67.39 &70.81 &67.88 &48.59 &54.23 &42.80  \\ 
    & RoLT &85.68 &85.43 &83.50 &80.92 &78.96 &73.02 &71.20 &66.53 &57.86 &48.98  \\
    & RoLT-DRW &86.24 &85.49 &84.11 &81.99 &80.05 &76.22 &74.92 &71.08 &63.61 &55.06 \\
    & RCAL &88.09 &86.46 &84.58 &83.43 &80.80 &78.60 &75.81 &72.76 &69.78 &65.05 \\ \cmidrule{2-12}
    & Our &\textbf{94.51} &\textbf{93.06} &\textbf{92.45} &\textbf{91.28} &\textbf{89.43} &\textbf{91.20} &\textbf{88.86} &\textbf{87.12} &\textbf{86.00} &\textbf{85.88} \\
\bottomrule
\toprule
Dataset    & Imbalance Ratio  &\multicolumn{5}{c}{10} &\multicolumn{5}{c}{100} \\ \midrule
\multirow{16}*{\rotatebox{90}{CIFAR-100}}    & Noise Rate    & 0.1 &0.2 & 0.3 &0.4 &0.5 & 0.1 &0.2 & 0.3 &0.4 &0.5   \\ \cmidrule{2-12}
    & ERM &48.54 &43.27 &37.43 &32.94 &26.24 &31.81 &26.21 &21.79 &17.91 &14.23  \\
    & ERM-DRW &50.38 &45.24 &39.02 &34.78 &28.50 &34.49 &28.67 &23.84 &19.47 &14.76  \\ \cmidrule{2-12}
    & LDAM &51.77 &48.14 &43.27 &36.66 &29.62 &34.77 &29.70 &25.04 &19.72 &14.19    \\
    & LDAM-DRW &54.01 &50.44 &45.11 &39.35 &32.24 &37.24 &32.27 &27.55 &21.22 &15.21 \\ 
    & CRT &49.13 &42.56 &37.80 &32.18 &25.55 &32.25 &26.31 &21.48 &20.62 &16.01 \\
    & NCM &50.76 &45.15 &41.31 &35.41 &29.34 &34.89 &29.45 &24.74 &21.84 &16.77     \\
    & MiSLAS &57.72 &53.67 &50.04 &46.05 &40.63 &41.02 &37.40 &32.84 &26.95 &21.84 \\ \cmidrule{2-12}
    & Co-teaching &45.61 &41.33 &36.14 &32.08 &25.33 &30.55 &25.67 &22.01 &16.20 &13.45 \\
    & CDR &47.02 &40.64 &35.37 &30.93 &24.91 &27.20 &25.46 &21.98 &17.33 &13.64 \\
    & Sel-CL+ &55.68 &53.52 &50.92 &47.57 &44.86 &37.45 &36.79 &35.09 &31.96 &28.59  \\ \cmidrule{2-12}
    & HAR-DRW &51.04 &46.24 &41.23 &37.35 &31.30 &33.21 &26.29 &22.57 &18.98 &14.78  \\ 
    & RoLT &54.11 &51.00 &47.42 &44.63 &38.64 &35.21 &30.97 &27.60 &24.73 &20.14  \\
    & RoLT-DRW &55.37 &52.41 &49.31 &46.34 &40.88 &37.60 &32.68 &30.22 &26.58 &21.05 \\
    & RCAL &57.50 &54.85 &51.66 &48.91 &44.36 &41.68 &39.85 &36.57 &33.36 &30.26 \\ \cmidrule{2-12}
    & Our &\textbf{77.90} &\textbf{75.32} &\textbf{74.66} &\textbf{72.54} &\textbf{70.53} &\textbf{74.05} &\textbf{71.47} &\textbf{71.37} &\textbf{69.03} &\textbf{66.31} \\
\bottomrule
\end{tabular}
% \end{sc}
\end{small}
\end{center}
\vskip -0.1in
% \end{table}
%%%%%%%%%%%%%%%%%%%%%%%%%%%%%%%%%%%%%%
%%%%%%%%%%%%%%%%%%%%%%%%%%%%%%%%%%%%%%

\section{Comparison between BAlanced Noise-tolerant Cross entropy (BANC)  and Symmetric Cross Entropy (SCE)}
\label{appendix:sce}

As discussed in Sec.~\ref{section:contrastive}, our proposed BAlanced Noise-tolerant Cross entropy (BANC) is inspired by Symmetric Cross Entropy (SCE). We introduce a linear function $c\,(1-\bar{y}_i)$ in the symmetric term, and through the scaling coefficient $c$, we adjust the penalty of the loss for mispredictions, enhancing tolerance to noisy labels and allowing the model to learn each class more balanced. In this section, we compare the noise resistance and balancing capabilities of BANC and Symmetric Cross Entropy (SCE) losses through experiments on the simulated CIFAR-100 dataset. Table~\ref{table:sce} presents their performance under different conditions: imbalance ratios of 10 and 100, with symmetric noise rates of 0.4 and 0.6. We also analyze their performance in three class sub-groups: many-shot classes, medium-shot classes, and few-shot classes.

%%%%%%%%%%%%%%%%%%%%%%%%%%%%%%%%%%%%%%
%%%%%%%%%%%%%%%%%%%%%%%%%%%%%%%%%%%%%%
\makeatletter\def\@captype{table}\makeatother
% \begin{table}[t]
\caption{Testing accuracy (\%) on simulated CIFAR-100. Bold shows the highest score.}
\label{table:sce}
\vskip 0.15in
\begin{center}
\begin{small}
% \begin{sc}
\begin{tabular}{c|cccc|cccc}
\toprule
Imbalance Ratio &\multicolumn{8}{c}{10} \\  \midrule
Noise Rate (\textbf{Sym.})  &\multicolumn{4}{c}{0.4} &\multicolumn{4}{|c}{0.6} \\ \midrule
Loss    & Many &Medium &Few &All  & Many &Medium &Few &All   \\ \midrule
SCE    & 79.88 &62.33 &\textbf{100.00} &73.64  & 78.32 &51.54 &\textbf{100.00} &69.75  \\ 
BANC & \textbf{80.68} &\textbf{62.52} &\textbf{100.00} &\textbf{74.30}  & \textbf{78.45} &\textbf{51.65} &\textbf{100.00} &\textbf{69.98}  \\
\bottomrule
\toprule
Imbalance Ratio &\multicolumn{8}{c}{100} \\  \midrule
Noise Rate (\textbf{Sym.})  &\multicolumn{4}{c}{0.4} &\multicolumn{4}{|c}{0.6} \\ \midrule
Loss    & Many &Medium &Few &All  & Many &Medium &Few &All   \\ \midrule
SCE    & \textbf{91.82} &76.26 &36.92 &71.17  & \textbf{92.87} &71.93 &22.97 &65.87  \\ 
BANC & 90.77 &\textbf{76.38} &\textbf{41.28} &\textbf{72.03}  & 92.31 &\textbf{72.74} &\textbf{23.84} &\textbf{66.64}  \\
\bottomrule
\end{tabular}
% \end{sc}
\end{small}
\end{center}
\vskip -0.1in
% \end{table}
%%%%%%%%%%%%%%%%%%%%%%%%%%%%%%%%%%%%%%
%%%%%%%%%%%%%%%%%%%%%%%%%%%%%%%%%%%%%%

When the imbalance ratio is 10, BANC outperforms SCE overall under both noise rates. When the imbalance ratio is 100, BANC performs significantly better than SCE in medium-shot and few-shot classes, resulting in higher overall accuracy. This demonstrates that BANC not only exhibits superior noise resistance but also effectively balances learning across different classes.

%%%%%%%%%%%%%%%%%%%%%%%%%%%%%%%%%%%%%%
%%%%%%%%%%%%%%%%%%%%%%%%%%%%%%%%%%%%%%
%%%%%%%%%%%%%%%%%%%%%%%%%%%%%%%%%%%%%%
\section{Influence of the scaling coefficient $\bm{c}$} \label{appendix:c}

\begin{figure}[ht]
\vskip 0.2in
\begin{center}
\centerline{\includegraphics[width=0.5\textwidth, height=0.3\textwidth]{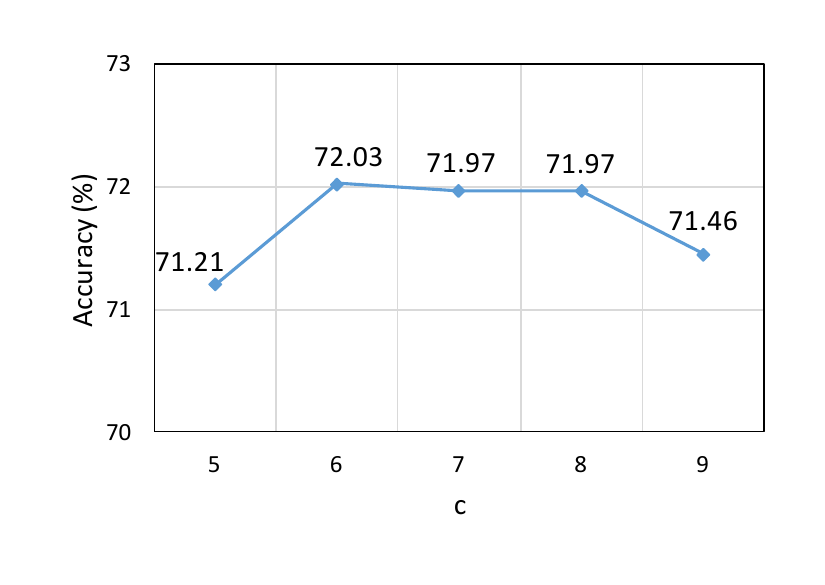}}
\caption{Impact of the scaling coefficient $c$ on classification accuracy is examined, with the optimal result observed when $c=6$. Note that this result solely represents predictions from the first stage and does not incorporate label refurbishment and multi-expert ensemble learning.}
\label{fig:scaling}
\end{center}
\vskip -0.2in
\end{figure}

The scaling coefficient $c$ in the BAlanced Noise-tolerant Cross entropy (BANC) loss controls the penalty for mis-predictions, influencing BANC's ability to handle noisy labels under class imbalance. We conducted an experiment to investigate the most appropriate value of $c$ on the simulated CIFAR100 dataset, with  symmetric noise rate 0.4 and imbalance ratio 100. Fig.~\ref{fig:scaling} shows that the best performance is achieved when $c = 6$.

%%%%%%%%%%%%%%%%%%%%%%%%%%%%%%%%%%%%%%
%%%%%%%%%%%%%%%%%%%%%%%%%%%%%%%%%%%%%%
% \makeatletter\def\@captype{table}\makeatother
% % \begin{table}[t]
% \caption{Test accuracy (\%) of many/medium/few classes on the simulated CIFAR-100 with an asymmetric noise rate of 0.4 and an imbalance ratio of 100. The best results are in bold.}
% \label{table:c}
% \vskip 0.15in
% \begin{center}
% \begin{small}
% % \begin{sc}
% \begin{tabular}{c|ccccc}
% \toprule
% The scaling coefficient $c$   &5     &6   &7  & 8   &9   \\ \midrule
% Accuracy    &65.97    &\textbf{67.01}    &66.68   &64.38  &65.93   \\ 
% \bottomrule
% \end{tabular}
% % \end{sc}
% \end{small}
% \end{center}
% \vskip -0.1in
% % \end{table}
%%%%%%%%%%%%%%%%%%%%%%%%%%%%%%%%%%%%%%
%%%%%%%%%%%%%%%%%%%%%%%%%%%%%%%%%%%%%%
%%%%%%%%%%%%%%%%%%%%%%%%%%%%%%%%%%%%%%
%%%%%%%%%%%%%%%%%%%%%%%%%%%%%%%%%%%%%%
%%%%%%%%%%%%%%%%%%%%%%%%%%%%%%%%%%%%%%
%%%%%%%%%%%%%%%%%%%%%%%%%%%%%%%%%%%%%%

\section{Influence of the hyperparameter $\bm{\alpha}$} \label{appendix:alpha}

\begin{figure}[ht]
\vskip 0.2in
\begin{center}
\centerline{\includegraphics[width=0.65\textwidth, height=0.35\textwidth]{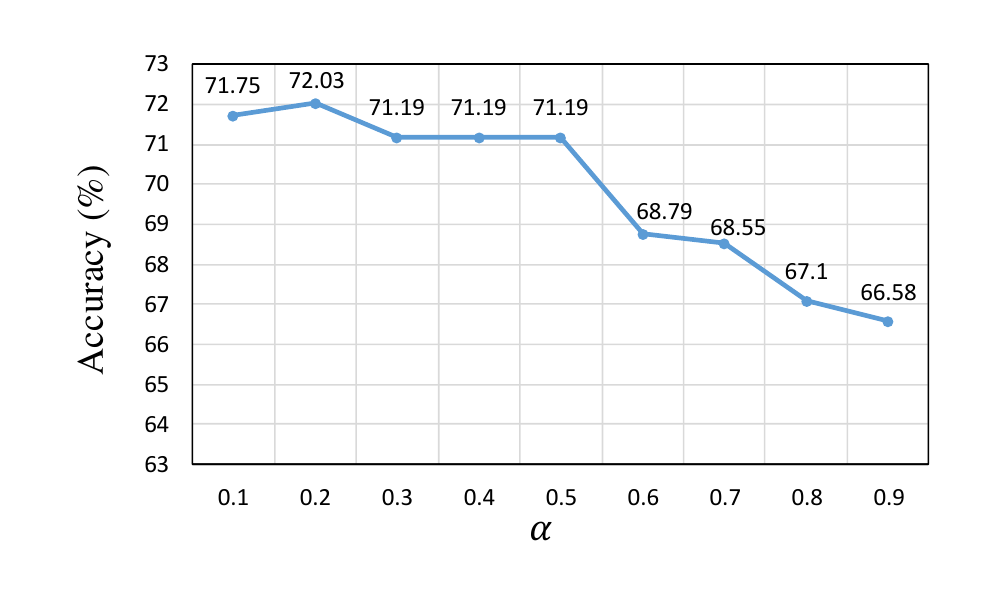}}
\caption{The effect of hyperparameter $\alpha$ on classification accuracy is investigated, and the optimal result is observed when $\alpha = 0.2$. Note that this outcome solely reflects predictions from the first stage and does not encompass label refurbishment and multi-expert ensemble learning.}
\label{fig:alpha}
\end{center}
\vskip -0.2in
\end{figure}

The hyperparameter $alpha$ controls the contribution of contrastive loss $\mathcal{L}_{con}$ and the BANC loss $\mathcal{L}_{BANC}$. We conducted an experiment here to investigate the most suitable value of $\alpha$ on the simulated CIFAR100 dataset with symmetric noise rate 0.4 and imbalance ratio 100. Fig.~\ref{fig:alpha} shows that the best performance is achieved when $\alpha = 0.2$.

%%%%%%%%%%%%%%%%%%%%%%%%%%%%%%%%%%%%%%
%%%%%%%%%%%%%%%%%%%%%%%%%%%%%%%%%%%%%%
%%%%%%%%%%%%%%%%%%%%%%%%%%%%%%%%%%%%%%
%%%%%%%%%%%%%%%%%%%%%%%%%%%%%%%%%%%%%%

\section{Illustration of the Rarity Score $\bm{\gamma_i}$ in Eq. (\ref{eq:ri})}
\label{appendix:ri}
The rarity score $\gamma_i$ is estimated as a function inversely proportional to the proportion of class $\bar{y}_i$ in the dataset. Let $n_{\bar{y}_i}$ denote the number of samples belonging to class $\bar{y}_i$, and let the proportion of class $\bar{y}_i$  be $h_i=\frac{n_{\bar{y}_i}}{N}$. We uses a normal distribution function with zero mean to model $\gamma_i$: when $h_i$ is close to zero, $\gamma_i$ approaches the maximum value 1, and when $h_i$ is greater than 0.5, $\gamma_i$ quickly decays to zero.  From Fig. A3, the variance $\sigma$ of Eq.(\ref{eq:ri}) is estimated as 0.2.  
%%%%%%%%%%%%%%%%%%%%%%%%%%%%%%%%%%%%%%
%%%%%%%%%%%%%%%%%%%%%%%%%%%%%%%%%%%%%%
\begin{figure}[ht]
\vskip 0.2in
\begin{center}
\centerline{\includegraphics[width=0.5\textwidth]{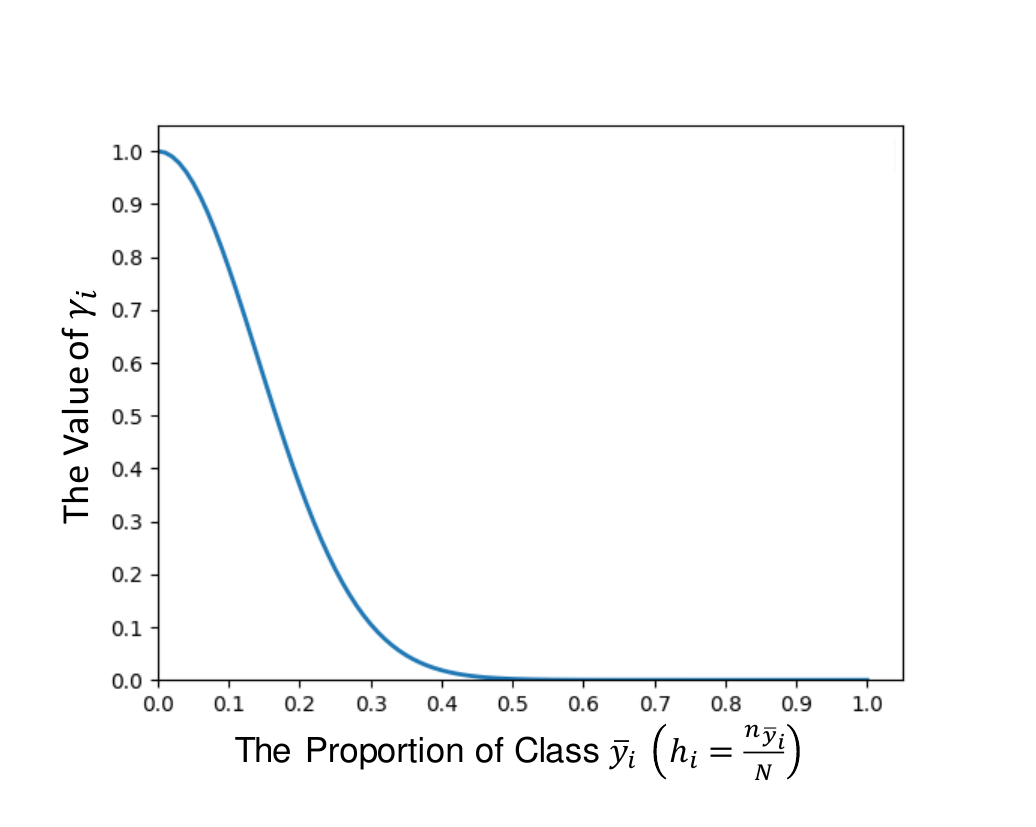}}
\caption{Plot of the rarity score $\gamma_i$, which is inversely proportional to the number of samples in class $k$.}
\label{fig:ri}
\end{center}
\vskip -0.2in
\end{figure}
%%%%%%%%%%%%%%%%%%%%%%%%%%%%%%%%%%%%%%
%%%%%%%%%%%%%%%%%%%%%%%%%%%%%%%%%%%%%%
\end{document}